\documentclass[lettersize,journal]{IEEEtran}
\usepackage{algorithm}
\usepackage{array}
\usepackage[caption=false,font=normalsize,labelfont=sf,textfont=sf]{subfig}
\usepackage{textcomp}
\usepackage{stfloats}
\usepackage{url}
\usepackage{verbatim}
\usepackage{graphicx}
\usepackage{cite}
\usepackage{amsmath,amssymb,amsfonts}
\usepackage{xcolor}
\usepackage{xurl} 
\usepackage[colorlinks=false]{hyperref} 
\usepackage{algpseudocode}
\usepackage{booktabs}   
\usepackage{multirow} 
\usepackage{tabularx}
\usepackage{orcidlink}  
\usepackage{tcolorbox}

\begin{document}

\title{When Cognitive Graphs Meet LLMs: BDEI Cognitive Pathways for Panic Emotional Arousal Prediction}


\markboth{Journal of \LaTeX\ Class Files,~Vol.~14, No.~8, August~2021}%
{Shell \MakeLowercase{\textit{et al.}}: A Sample Article Using IEEEtran.cls for IEEE Journals}

\IEEEpubid{0000--0000/00\$00.00~\copyright~2021 IEEE}

\author{Mengzhu~Liu~\orcidlink{0009-0005-3490-4774}\textsuperscript{†}, Long~Qin~\orcidlink{0000-0003-1245-6622}\textsuperscript{†}, Chuan~Ai~\orcidlink{0000-0002-9641-6267}, Zhengqiu~Zhu~\orcidlink{0000-0002-5805-834X}, Hongru~Liang~\orcidlink{0000-0002-4776-4505}, Fangfang~Li~\orcidlink{0000-0001-7829-1539}, Chen~Gao~\orcidlink{0000-0002-7561-5646}, Yong~Li~\orcidlink{0000-0001-5617-1659}, Xin~Lu~\orcidlink{0000-0002-3547-6493}, and Quanjun~Yin~\orcidlink{0000-0002-1633-174X}%
\thanks{
This work was supported in part by the National Natural Science Foundation of China under Grant 72604448.
\textit{\textsuperscript{†}Equal contribution}}
\thanks{Mengzhu Liu, Long Qin, Chuan Ai, Zhengqiu Zhu, Xin Lu, and Quanjun Yin  are with National Key Laboratory of Digital Intelligent Modeling and Simulation, National University of Defense Technology, Changsha 410073, Hunan Province, China. (e-mail: liumengzhu2001@nudt.edu.cn; qinlong@nudt.edu.cn; aichuan@nudt.edu.cn; zhuzhengqiu12@nudt.edu.cn; xin.lu.lab@outlook.com; yin\_quanjun@163.com).}
\thanks{Hongru Liang is with the College of Computer Science, Sichuan University, Chengdu 610065, China (e-mail: lianghongru@scu.edu.cn).}
\thanks{Fangfang Li is with the School of Computer Science and Engineering, Central South University, Changsha 410083, China (e-mail: lifangfang@csu.edu.cn).}
\thanks{Chen Gao and Yong Li are with the Beijing National Research Center for Information Science and Technology (BNRist), Tsinghua University, Beijing100084, China (e-mail: chgao96@tsinghua.edu.cn; liyong07@tsinghua.edu.cn).}
\thanks{The code and dataset are available at https://github.com/supersonic0919/PanicCognitivePath.}

}

\maketitle

\begin{abstract}

Predicting the timing of individual and collective panic emotional arousal before manifestation is essential for timely emergency intervention.  Existing methods incorporate cognitive elements but none of them model emotion in the generative direction of the arousal process, leaving arousal timing undetermined. We argue that grounding prediction in appraisal emotion theory is necessary because it models this process explicitly in its natural generative direction, but three problems must be solved. (1) Appraisal theory posits that emotion arises from simultaneous evaluation across multiple threat dimensions, yet no prior work fuses these inputs into risk perception; (2) Existing models are trained in the opposite, behavior-bridged direction, recovering emotion merely as a post-hoc correlate of behavior; (3) Approaches that adopt LLMs as the primary decision-maker yet overlook the fragility and hallucination-proneness of their outputs. We introduce PanicCognitivePath (PCP) to address all three. A Psychological Safety Distance (PSD) model, grounded in psychological distance theory, maps four-domain signals (physical, social, cognitive, and informational) into a unified risk metric that gates entry to cognitive reasoning. An explicit Emotion node grounded in appraisal emotion theory is introduced into BDI, forming a novel Belief-Desire-Emotion-Intention (BDEI) pathway that couples threat appraisal directly to emotional arousal. Inverting the conventional LLM-as-decision-maker paradigm, PCP confines the LLM to parameter estimation for the Belief-to-Desire transition, restricting hallucinations to a single step and curbing their accumulation across steps. Experiments on Hurricane Sandy show PCP improves individual prediction accuracy by \textbf{10.68\%}\footnote{All percentage symbols (\%) in this paper denote percentage points (pp.) unless otherwise specified.} over baselines, reduces peak count error to \textbf{7.07\%}. 

\end{abstract}


\begin{figure*}[!t]
\centering
\includegraphics[width=\textwidth]{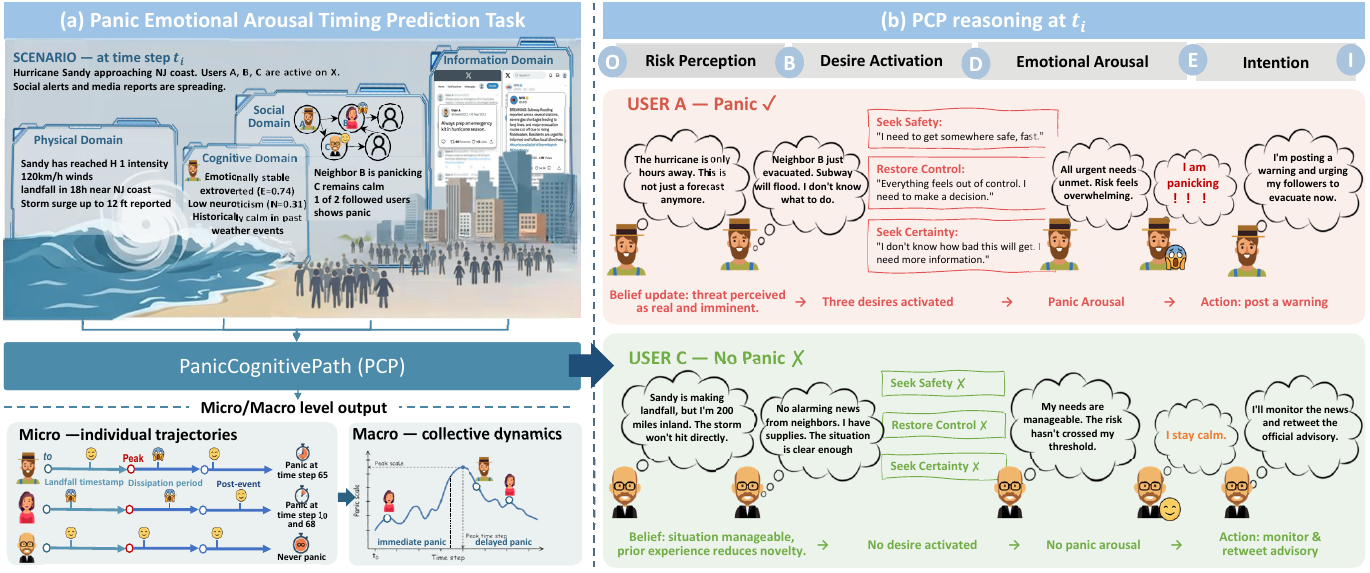}
\caption{\textbf{Multi‑domain coupling for panic arousal timing and macro outbreak prediction.} \textbf{(a) Task formulation.} At the time step $t_i$, physical, social, cognitive and information inputs drive PCP reasoning to predict each individual's panic arousal timing and the macro peak panic scale and its time step. \textbf{(b) PCP reasoning at step $t_i$.} User A activates all three desires and panics; User C perceives lower risk, activates no desire, and stays calm.}
\label{fig:pcp_framework_in_hurricane_disaster}
\end{figure*}


\section{Introduction}

Panic emotional arousal (abbreviated as panic arousal thereafter) during emergencies can transform affected populations into secondary hazard sources, as its propagation often exceed the physical damage itself~\cite{sheikhi2025looting}. For instance, after the 2010 Chilean magnitude-8.8 earthquake, panic and associated social disorder compounded the original disaster~\cite{garfin2014exposure}. This motivates the task of panic arousal prediction: as illustrated in Figure~\ref{fig:pcp_framework_in_hurricane_disaster}, given pre-disaster user characteristics and real-time emergency signals, can we predict each individual's
panic arousal timing and the collective peak scale with its time step before panic manifests? Such prediction enables semi-prospective intervention: cognitive profiles are constructed offline before disaster onset, and real-time prediction combines them with live emergency signals.

Existing approaches to panic arousal timing prediction include differential equation models and agent-based simulations, which rely on averaged population assumptions or predefined rules~\cite{wu2025novel}. Further, some integrate the Belief-Desire-Intention (BDI) architecture into agent modeling~\cite{adam2016bdi}, introducing structured reasoning from beliefs and desires to intentions. More recently, LLM-driven generative agents offer contextual reasoning~\cite{liu2025pychoagent}, with improvements via knowledge graphs~\cite{wu2024coke}, retrieval-augmented generation~\cite{ye2025emotiongraphrag}, and chain-of-thought prompting. However, these approaches neglect to model the emotional arousal process. The transition from risk perception to arousal remains unmodeled, leaving predictions of arousal timing without psychological fidelity.\IEEEpubidadjcol

We argue that grounding panic arousal timing prediction in appraisal emotion theory~\cite{ong2019computational} is necessary because it explicitly models how emotional arousal arises from cognitive appraisal of events. To achieve this, three challenges must be addressed. (1) Appraisal involves stimuli from multiple domains, yet integrating physical, social, cognitive, and informational signals into a unified representation remains underexplored. Without such fusion, the model captures only partial risk signals, falling short of the comprehensive threat evaluation that appraisal theory requires. (2) Existing cognitive models infer emotions indirectly from behaviors rather than explicitly modeling their formation process. This behavior‑bridged modeling direction gives rise to two critical issues for arousal‑timing prediction. First, the model does not learn the generative path from appraisal to arousal and thus provides no account of when emotional arousal is produced. Second, when a panic state is inferred at a certain time step, and its arousal estimates are not traceable to identifiable appraisal steps. Even with a complete cognitive pathway, parameterizing individual-level transitions from threat perception to desire activation requires sensitivity to personal heterogeneity. (3) Given their generalizable cognitive reasoning, current approaches adopt LLMs as the primary decision-maker, yet overlook the fragility and hallucination-proneness of their outputs. Once the LLM produces a hallucinated emotional response at one step, this error propagates and amplifies across subsequent simulation steps. Without circumscribing the LLM within a structured pathway, outputs become unreliable\cite{xue2026some}.


To ground these challenges, consider two residents during Hurricane Sandy (Figure~\ref{fig:pcp_framework_in_hurricane_disaster}). Resident A, near the storm center with panicking neighbors and high neuroticism, panics immediately upon landfall; Resident C, in the same impact zone but with calm social ties, stronger emotional regulation, and limited information access, remains non-panic throughout the event. Both face equivalent physical threat, yet their panic outcomes diverge entirely due to differences in social influence, cognitive traits, and information exposure---factors that no single-domain risk model can jointly capture. Moreover, explaining why C stays calm while A panics requires tracing the full cognitive process from threat perception through belief updating to emotional arousal, which behavior-bridged inference cannot represent the generative appraisal pathway. Finally, an unconstrained LLM asked to predict either resident step-by-step may produce plausible but temporally inconsistent trajectories, as nothing in its prompt enforces stable cognitive transitions across time steps. These limitations call for a framework that couples multi-domain signals, models the cognitive pathway explicitly, and constrains the LLM within a structured graph.

To address these challenges, we propose PanicCognitivePath (PCP), a BDEI (Belief-Desire-Emotion-Intention) cognitive path framework. (1) To address the risk perception representation challenge, we construct a psychological safety distance model (PSD) for mapping heterogeneous four-domain signals into a unified risk metric, which determines pathway entry via a safety threshold. (2) To address the pathway challenge, we introduce an explicit Emotion node grounded in appraisal emotion theory into BDI, forming a BDEI pathway. This node couples threat appraisal directly to emotional arousal, making the threat-to-arousal transition a traceable cognitive step. (3) To address the reliability challenge, we constrain the LLM to parameter inference for only the Belief-to-Desire transition within the BDEI pathway. Rigid rule-based thresholds cannot capture how individual traits and situational contexts jointly shape desire activation, making the LLM's contextual reasoning necessary. However, placing the LLM at the center of decision-making allows hallucinations to propagate across steps. By confining the LLM to a well-defined subtask within a structured cognitive graph, hallucinations are isolated to a single step, mitigating cross-step error accumulation. 

Our contributions are summarized as follows:
\begin{itemize}
\item \textbf{Conceptually,} we ground panic emotional arousal timing prediction in appraisal emotion theory, recognizing that existing methods model emotion in the behavior-bridged direction and therefore lack a generative account of the arousal process. We identify three sequential barriers to achieving this grounding, reframing the task from behavior prediction to psychologically grounded emotion timing prediction.
\item \textbf{Methodologically,} we propose PCP, integrating a psychological safety distance model for unified risk perception, a BDEI pathway with an explicit Emotion node, and a constrained LLM role that performs only parameter inference. The structured cognitive graph governs all state transitions, making the panic arousal process traceable and limiting LLM hallucination propagation to single steps.
\item \textbf{Experimentally,} we curate a Hurricane Sandy dataset with three-class individual panic arousal labels (none/immediate/delayed panic) and evaluate PCP against 12 baselines. PCP improves individual arousal timing accuracy by 10.68 percentage points, reduces peak panic count error to 7.07\%. These results demonstrate that explicit modeling of the emotional arousal process enables both accurate individual prediction and faithful collective dynamics.

\end{itemize}

\section{Related Work}
\subsection{Traditional Simulation Models of Emotion Propagation}
Social simulation under emergencies relies on dynamic models and ABMs. Epidemic-derived models (e.g., SIS/SIR, SLIRS\cite{wang2016computational}, SI-SEIR\cite{wang2025modeling}) treat emotions as pathogen transmission. Although efficient at the macro level, they reduce arousal to state probabilities and miss individual cognition and multi-domain coupling\cite{chu2024emotional}. ABMs and social force models allow heterogeneity but use rigid rules that cannot model implicit perception-to-arousal pathways\cite{zhao2024improved}. Both share two limitations: 1) reducing arousal to probabilistic transitions while ignoring how individuals cognitively appraise threats; 2) relying on single-dimensional data, which fails to capture how physical hazards, social ties, and information exposure jointly shape panic, leaving the model blind to cross-domain interactions that drive real-world arousal.

\subsection{LLM-Driven Social Simulation}
Recent years have seen a growing trend of using LLM-based agents for social simulation in domains such as transportation, economics, and public opinion. Park et al. showed that LLM-driven agents replicate social behaviors via reflection and planning\cite{park2023generative}, with recent platforms (AgentSociety\cite{piao2025agentsociety}, LAID\cite{hu2024llm}, YuLan-OneSim\cite{wang2025yulan}) further enhancing fidelity through natural language interaction. Researchers have also adopted GraphRAG to ground agent reasoning in structured knowledge\cite{wu2024coke,shen2025coe}. However, applying LLMs to panic simulation remains problematic. Current LLM-driven approaches place the model at the center of reasoning, leading to unstable, prompt-sensitive outputs and accumulated hallucination across simulation steps. Moreover, purely LLM-driven agents lack computable coupling with external multi-domain data, and their opaque reasoning chains cannot produce traceable arousal predictions required for risk assessment\cite{ye2025emotiongraphrag}. Complementary evaluation efforts probe LLMs' emotional competence. Both studies \cite{NEURIPS2024_b0049c3f,Ala2025Aware} report systematic misalignment between LLM-generated affect and human affect, which reinforces our design decision to confine the LLM to single-step parameter estimation within a structured cognitive pathway rather than granting it end-to-end emotional decision-making.

\subsection{Cognitive Architectures and Psychological Foundations for Agent Emotion Modeling}
Emotion propagation theory holds that affective states spread via primitive mimicry and feedback\cite{hatfield1993emotional} across physical and online channels, precisely the multi-domain coupling that remains challenging to operationalize in computational models. Psychologically, appraisal theories (e.g., OCC \cite{ortony2022cognitive, wang2024logic}) posit that emotions arise from structured event appraisal, providing a foundation for emotion simulation. Yet existing implementations oversimplify elicitation pathways, neglect intuition-based emotions and psychological safety distance, and cannot dynamically integrate multimodal cues for panic arousal timing prediction\cite{zhong2026modeling}. Furthermore, the BDI cognitive architecture formalizes rational agent behavior in multi-agent systems \cite{rao1995bdi, de2020bdi}, although, in its canonical form, it contains no affective component, although a substantial affective computing literature has enriched BDI-style architectures with appraisal-based emotion generation\cite{popescu2014,jiang2007ebdi}. However, these models serve the believability of synthetic agents: emotion is generated from hand-crafted appraisal rules to modulate action selection, and arousal timing is never the prediction target.

\section{Problem Definition}

\subsection{Definition of Panic Arousal Timing}

Panic propagation in crises exhibits salient temporal characteristics that distinguish it from panic emotion recognition. While recognition tasks identify whether an individual is panicking at a given moment, propagation demands predicting when and how quickly arousal spreads across a population over time. To quantify this dynamic, we define it at the level of micro-individual psychological processing.


At the individual level, the panic arousal time is the interval from an
individual $a_i$'s first exposure to an information stimulus of emergency
event $E$ to the moment when panic arousal becomes manifest. This process
encompasses not only the physical arrival time of the information, but
also the psychological time required for the individual to appraise the
threat through cognitive appraisal based on their psychological schema.
Formally, let $t_0$ denote the onset of the emergency event, and let
$s_i(t)\in\{0,1\}$ denote the binary panic arousal state of
individual $a_i$ at time step $t$, where $s_i(t)=1$ indicates that panic
arousal is manifest at step $t$, and $s_i(t)=0$ otherwise. The panic
arousal time $\tau_i$ of individual $a_i$ is then defined as follows:

\begin{equation}
\tau_i \;=\; \min \{\, t \ge t_0 \;:\; s_i(t) = 1 \,\},
\label{eq:arousal-time}
\end{equation}

If the individual never manifests panic arousal throughout the simulation cycle, $\tau_i = \infty$.


At the macro level, let $N(t)=|\{\,a_i : s_i(t)=1\,\}|$ denote the number
of panicking individuals at time step $t$. The collective eruption
dynamics are characterized by the peak panic count
$N^{*}=\max_{t} N(t)$ and the peak time step
$t^{*}=\arg\max_{t} N(t)$.

\subsection{From Behavior-Conditioned Inference to Emotion-Generative Modeling}

Most existing approaches learn emotion in a behavior-bridged direction: they first fit a mapping from inputs to observable behavior and then recover emotional states from that behavior, obtaining emotion as a two-step backward correlate rather than as a state produced by the appraisal process. We formalize the two directions below. Throughout, we emphasize that this is a distinction in modeling direction and target quantity. The distinction concerns what a predictive model represents, not whether affect is observable.

\subsubsection{Behavior-Bridged Direction}

Given the single-domain prior information $D_{\text{phy}}/D_{\text{soc}}/D_{\text{cog}}/D_{\text{info}}$ at time $t_0$ of emergency $E$, and the emotional labels $Y = \{y_i\}_{i=1}^{n}$ ($y_i \in \{0,1\}$), the task solves for a composite mapping $\Phi$:

\begin{equation}
\Phi: (D, Y \mid E, t_0) \xrightarrow{f_{\text{beh}}} B \xrightarrow{f_{\text{inv}}} \{(\hat{y}_i, \hat{\tau}_i)\}_{i=1}^{n},
\end{equation}
where $f_{\text{beh}}$ maps the single-domain input $D$ to behavioral traces $B$, and $f_{\text{inv}}$ inversely recovers the panic state $\hat{y}_i \in \{0,1\}$ and arousal time $\hat{\tau}_i \in \mathbb{R}^+$ (if $\hat{y}_i = 1$; else undefined) from $B$. This paradigm has two limitations. First, it estimates a conditional association between behavior and emotion, so two distinct appraisal pathways that yield similar behavior remain indistinguishable, and the model carries no internal representation of how arousal is produced. Second, single-domain input cannot capture how heterogeneous factors jointly drive panic arousal.

\subsubsection{Emotion-Generative Direction}

Given the multi-domain prior information $D_{\text{phy}}, D_{\text{soc}}, D_{\text{cog}}, D_{\text{info}}$ at time $t_0$ of emergency $E$, where $D_{\text{cog}}$ and $D_{\text{info}}$ are pre-disaster user features mined from historical data, while $D_{\text{phy}}$ and $D_{\text{soc}}$ provide real-time disaster and network states. The task directly models the forward cognitive appraisal process from stimulus perception to emotional arousal. Formally:

\begin{equation}
\Psi: (D_{\text{phy}}, D_{\text{soc}}, D_{\text{cog}}, D_{\text{info}} \mid E, t_0) \xrightarrow{\text{appraise}} \{(\hat{y}_i, \hat{\tau}_i)\}_{i=1}^{n},
\end{equation}

Unlike $\Phi$, $\Psi$ is a structured cognitive pathway grounded in appraisal emotion theory, and its object of learning is $P(\text{arousal} \mid \text{multi-domain appraisal})$ rather than $P(\text{emotion} \mid \text{behavior})$. Each prediction is produced by a traceable appraisal process that proceeds from stimulus perception through belief formation, desire generation and emotional arousal to behavioral intention, making the arousal mechanism generative rather than inverted from behavioral outcomes.

The shift from $\Phi$ to $\Psi$ is thus a change of generative direction and of target quantity, not a rejection of behavior-based evidence. In $\Phi$, this bridge forces emotion to be a post-hoc recovered latent variable rather than a state produced through a traceable psychological process. In $\Psi$, emotion is generated forward as a first-class state via a structured appraisal pathway with multi-domain fusion as input. Panic arousal timing prediction demands a trustworthy account of how arousal occurs, which only a direct forward cognitive model can provide.

\begin{figure*}[htbp]
  \centering
  \includegraphics[width=\linewidth]{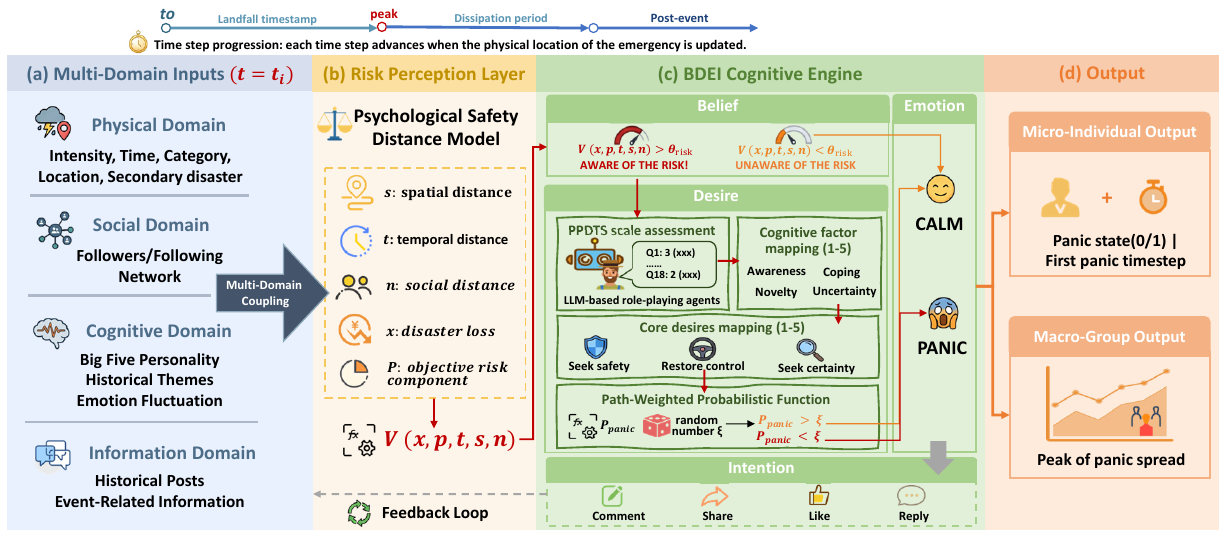}
  \caption{\textbf{Overview of the PCP framework.} Four-domain inputs are coupled via a PSD model into a unified risk metric; agents exceeding the safety threshold enter a BDEI pathway where the cognitive graph governs all transitions and the LLM serves only Belief-to-Desire parameter inference. Outputs include individual panic state and first arousal timestep (micro), and peak panic count with corresponding timestep (macro).}
  \label{fig:framework}
\end{figure*}

\section{PanicCognitivePath Method}
\subsection{Overview}

To overcome these limitations, we propose PCP, a multi-domain coupled cognitive framework that predicts panic arousal timing before overt manifestation. As illustrated in Fig.~\ref{fig:framework}, PCP transcends single-dimensional and pure LLM approaches by coupling physical, social, cognitive, and informational domains, covering four core elements: external stimulus, social amplification, cognitive processing, and information interaction. Specifically:

\begin{itemize}
    \item \textbf{Physical domain} captures emergency spatiotemporal features (intensity, time, location, disaster derivatives), characterizing objective disaster severity. 
    \item \textbf{Social domain} captures network topology (following/follower relations), quantifying social amplification of risk perception.
    \item \textbf{Cognitive domain} captures psychological heterogeneity (Big Five traits, topic preferences, emotional patterns), establishing differentiated response baselines to crises.
    \item \textbf{Information domain} integrates historical posts and event content, characterizing transmission paths and individual processing patterns.
\end{itemize}

At the core cognitive level, PCP extends the classical BDI framework\cite{rao1995bdi} with appraisal emotion theory to form a BDEI pathway for panic simulation. As illustrated in Fig.~\ref{fig:framework}, the pathway introduces an explicit Emotion node linking Belief/Desire to behavioral Intention. Events and mental states jointly trigger emotion, which drives expression and intentional behavior; these in turn update mental states and create new outcomes, forming a dynamic causal loop. By formalizing appraisal theory, BDEI decomposes implicit emotion arousal into node-traceable paths, addressing the affect gap of the canonical BDI architecture. Details are in Section B.

PCP orchestrates three components in a sequential pipeline with clear data-flow dependencies. The PSD model serves as the entry gateway, fusing four-domain signals into a unified risk metric that filters low-risk agents before cognitive processing, thereby reducing both false positives and computational cost. For agents that enter the pathway, the BDEI cognitive graph governs all state transitions through well-defined edge models (Sections C–E), while the LLM is confined to the Belief-to-Desire edge as a local parameter estimator for desire activation. This separation ensures that LLM outputs are structurally constrained: the cognitive graph controls how desire parameters propagate to emotion and behavior, localizing potential LLM errors to a single transition and mitigating the risk of cascading hallucination across steps. The framework outputs each agent's panic state per time step (binary: panic/calm) and first arousal time step, supporting both micro-individual prediction and macro-propagation analysis. Algorithm~\ref{alg:pcp} presents the single-step pseudocode for all agents.

\begin{algorithm}[t!]
\small
\caption{PCP Framework for Panic Elicitation Prediction}
\label{alg:pcp}
\begin{algorithmic}[1]
\Require Multi-domain inputs: physical events $\mathcal{P}$, social network $\mathcal{S}$, cognitive profiles $\mathcal{C}$, information history $\mathcal{I}$; time steps $T$; risk threshold $\theta_{\text{risk}}$
\Ensure Individual panic states $\{s_i(t)\}$ and first panic time $\tau_i$ for each agent $i$; macro propagation rate $R_{\text{macro}}$

\State \textbf{Initialize:} panic history $H \leftarrow \emptyset$; results list $\mathcal{R} \leftarrow \emptyset$
\For{$t = 1$ \textbf{to} $T$}
    \State Load current event state $e_t$ from $\mathcal{P}$
    \For{each agent $i$ in parallel}
        \State \textbf{Risk Perception:} $v_i \leftarrow \text{PsychSafetyDistance}(i, e_t, H, \mathcal{S})$ \\
\hspace*{2em} \Comment{Eq. (4)}
        \If{$v_i \le \theta_{\text{risk}}$}
            \State $s_i(t) \leftarrow 0$ \Comment{No risk perceived}
        \Else
            \State \textbf{LLM Role-Playing:} $r_i \leftarrow \text{LLM}_{\text{role}}(i, e_t, \mathcal{S}, \mathcal{C}, \mathcal{I})$ \\
\hspace*{2em} \Comment{18-question assessment}
            \State \textbf{Cognitive Appraisal:} $f_i \leftarrow \text{CalcFactors}(r_i)$ \\
\hspace*{2em} \Comment{Awareness, Novelty, Uncertainty, Coping}
            \State \textbf{Desire Activation:} $d_i \leftarrow \text{MapDesires}(f_i)$ \\
\hspace*{2em} \Comment{Three desires: safety, control, certainty}
            \State \textbf{Panic Generation:} $s_i(t) \sim \text{Bernoulli}(P_{\text{panic}})$ \\
\hspace*{2em} \Comment{Eq. (5)}
        \EndIf
        \State Update $H[i] \leftarrow H[i] \cup \{(t, s_i(t))\}$
        \State Record $\tau_i = \min\{t \mid s_i(t)=1\}$ if first panic
    \EndFor
    \State \textbf{Macro Propagation:} compute infection panic peak $I_{\text{peak}} = \max_i \sum_j s_j(t)$; $t_{\text{peak}} = \arg\max_t I_{\text{peak}}$
\EndFor
\State \textbf{return} $\{(\tau_i, s_i(t))\}_{i,t}$, $R_{\text{macro}}$
\end{algorithmic}
\end{algorithm}

\subsection{Construction of BDEI Cognitive Path Graph}
\begin{figure}[t!]
  \centering
  \includegraphics[width=\linewidth]{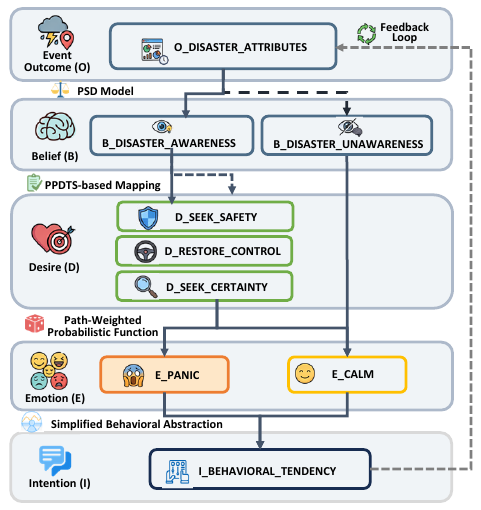}
  \caption{\textbf{BDEI cognitive pathway graph for panic arousal.} The graph formalizes appraisal emotion theory as a causal pathway of five node types (O, B, D, E, I) and sub-nodes, with directed edges annotated by computational models.}
  \label{fig:BDEI_cognitive_pathway}
\end{figure}

To transform panic from an implicit psychological process into a computable, traceable structured pathway, this section constructs the BDEI cognitive pathway graph for extreme emotion (Fig.~\ref{fig:BDEI_cognitive_pathway}). The graph defines five core nodes and their causal transitions in a directed format, fully characterizing the cognitive chain from disaster event perception to panic arousal.

\textbf{Node Design.} The graph has five node types. Event Outcome(O) encodes physical spatiotemporal features (disaster intensity, time, location). Belief (B) represents disaster perception (perceived/not perceived) with risk propensity. Desire(D) captures three psychological states: seeking safety, restoring control, and seeking certainty, each with quantitative multi-domain features. Emotion(E) outputs (panic/calm). Intention(I) covers behavioral tendencies for downstream propagation dynamics. Following appraisal theory, E lies between D and I as the emotional gate turning motivation into action readiness.

\textbf{Edge Design.} The graph defines five causal transitions: (1) Risk perception (O→B): Transfers event information to belief via psychological safety distance, determining whether the agent enters subsequent cognitive processing (Section C); (2) Desire arousal (B→D): An LLM role-playing agent evaluates the agent's psychological state by answering the Psychological Preparedness for Disaster Threats Scale (PPDTS)\cite{mclennan2020conceptualising}, and the resulting scores are mapped via fixed rules onto four appraisal factors (risk awareness, novelty, uncertainty, and coping efficacy), which are aggregated into the intensity and activation of three core desires (Section D); (3) Panic arousal (D→E): Uses a path-weighted probability function to output binary emotion (0=calm, 1=panic) (Section E); (4) Behavioral orientation (E→I): Captures the influence of emotion on behavioral intention. Targeting propagation speed prediction, we model emotional expression as propagation behavior (updating self state and influencing neighbors) instead of fine-grained actions; (5) Iterative feedback loop: Updated belief/emotion states feed into next-step risk perception and social distance, enabling continuous panic simulation.



\subsection{Calculation of Psychological Safety Distance}
Psychological Safety Distance (PSD) derives from psychological distance theory \cite{bullogh1912psychical,trope2010construal}, which views risk perception as the perceived distance to a risk stimulus. Multi-dimensional distance shapes risk cognition across temporal, spatial, social, and probability dimensions\cite{trope2007construal,smyth2007recipient}. Greater distance weakens perceived risk through a homogeneous discounting effect\cite{gattig2007judgmental}. We define PSD as the integrated multi-dimensional psychological distance from an individual to a sudden disaster. PSD is negatively correlated with psychological discounting and positively with subjective risk perception, serving as a core metric for initial disaster risk perception.

We compute PSD based on the Probabilistic Spatiotemporal Trade-off (PTST) model\cite{she2012psychological}, which integrates four psychological distance dimensions, decouples subjective discounting from objective risk, and is better suited for quantifying individual risk perception in sudden disasters. Tailored to hurricane-evacuation panic arousal, we adapt its input variables and parameters using PCP multi-domain data, as computed below.

\begin{equation}
    V(x,p,t,s,n)=e^{-(a\ln(|t|+1)+b\ln(s+1)+c\ln((1-n)+1))}\cdot p\cdot x,
\label{eq.4}
\end{equation}
where $V$ denotes the perceived risk level, with a larger value indicating stronger subjective risk perception of the hurricane. The discount function $e^{-(\cdot)}$ aggregates four psychological distance dimensions:

\begin{itemize}
    \item $t$: temporal distance, the time-step difference between the current step and disaster occurrence. We define $t$ as the signed time offset (in days) from disaster onset, with positive values indicating post-onset and negative values indicating the forecast window; $a$: temporal discounting weight.
    \item $s$: spatial distance, physical distance to the disaster epicenter; $b$: spatial discounting weight;
    \item $n$: social panic exposure, quantified as the proportion of panic-expressing agents; $c$: social discounting weight;
    \item $p \cdot x$: objective risk component, where $p=1$ for deterministic events and $x$ is the disaster loss, quantified by the hazard intensity level.
\end{itemize}

We set $a=b=c=1$, treating all distance dimensions equally. This follows the construal-level theory finding that the four dimensions share a common psychological meaning\cite{liberman2007psychological}. They produce similar effects on construal and risk perception, providing no theoretical basis for differential weighting without domain-specific priors.


\subsection{LLM-Assisted Inference for Belief-Desire Transition}

For precise B→D transition, we propose an LLM-assisted cognitive mapping method that converts vague disaster perception into quantifiable desire activation states. The method uses PPDTS as a psychometric instrument. An LLM role-playing agent completes the scale based on multi-domain user characteristics and real-time context, using the system-level and assessment prompts reproduced in Appendix~\ref{subsec:prompts}. Pre-defined rules then map responses to four cognitive factors, which are aggregated into three core desires with activation intensities.

\begin{table}[t!]
\caption{PPDTS item-to-factor and factor-to-desire mapping.
Reverse-scored items use \(s' = 5 - s\) before averaging;
factor scores are linearly remapped from the 1--4 raw scale to a
1--5 risk-perception scale.}
\label{tab:ppdts-mapping}
\centering
\footnotesize
\renewcommand{\arraystretch}{1.08}
\setlength{\tabcolsep}{2.5pt}
\begin{tabularx}{\columnwidth}{@{}c X l l@{}}
\toprule
\textbf{ID} & \textbf{Item (abbreviated)} & \textbf{Factor} & \textbf{Scoring} \\
\midrule
Q1  & Familiar with preparedness materials         & Novelty     & reverse \\
Q2  & Know how to prepare home                    & Uncertainty & reverse \\
Q3  & Know required preparedness measures         & Uncertainty & reverse \\
Q4  & Know what to look out for in emergency      & Awareness   & direct  \\
Q5  & Familiar with disaster warning system       & Novelty     & reverse \\
Q6  & Confident in actions during severe weather  & Uncertainty & reverse \\
Q7  & Can locate preparedness materials easily    & Uncertainty & reverse \\
Q8  & Knowledgeable about home impact             & Awareness   & direct  \\
Q9  & Know warning vs watch distinction           & Awareness   & direct  \\
Q10 & Familiar with weather signs of approach     & Novelty     & reverse \\
Q11 & Can manage feelings in difficult situations & Coping      & direct  \\
Q12 & Cope with anxiety and fear                  & Coping      & direct  \\
Q13 & Stay cool and calm in most difficulties     & Coping      & direct  \\
Q14 & Confident in dealing with stressful events  & Coping      & direct  \\
Q15 & Talk myself through challenging situations  & Coping      & direct  \\
Q16 & Know how to manage response                 & Coping      & direct  \\
Q17 & Know strategies to calm self                & Coping      & direct  \\
Q18 & Idea of how I would respond in emergency    & Coping      & direct  \\
\midrule
\multicolumn{4}{@{}l@{}}{\textbf{Factor $\to$ Desire mapping:}} \\
\multicolumn{4}{@{}l@{}}{\quad Safety desire: $I_S=\mathrm{avg}(awareness, novelty, uncertainty, coping)$} \\
\multicolumn{4}{@{}l@{}}{\quad Control desire: $I_C=\mathrm{avg}(awareness, coping)$} \\
\multicolumn{4}{@{}l@{}}{\quad Certainty desire: $I_{Ce}=\mathrm{avg}(novelty, uncertainty)$} \\
\bottomrule
\end{tabularx}
\end{table}

In implementation, an LLM agent builds a personalized profile from cognitive and information domains, integrates real-time physical disaster state and social panic ratio, and follows a 1–4 scale to complete the 18 PPDTS items. Responses reflect the user's true cognitive state under given constraints. Predefined mapping rules and linear transformation convert raw scores to a 1–5 range, yielding four cognitive factor scores: risk awareness, risk novelty, risk uncertainty, and coping efficacy. These factors represent cognitive depth, experience relevance, situational ambiguity, and self-efficacy, jointly forming the core cognitive appraisal basis for emotion arousal\cite{lambert1970psychological}. Based on disaster psychology, these factors are weighted to compute activation intensities for three core desires: seek safety (all four factors), restore control (awareness and coping), and seek certainty (uncertainty and novelty). Table~\ref{tab:ppdts-mapping} summarizes the 18-item PPDTS questionnaire, the four-factor structure used in our model, and the factor-to-desire mapping. The full item wording is reproduced verbatim in Appendix~\ref{subsec:ppdts_ques}. A desire activates when its intensity exceeds a threshold. This mechanism, grounded in cognitive appraisal theory and need for cognitive closure\cite{kruglanski1996motivated}, translates discrete appraisal dimensions into motivationally significant desire states.

This design converts implicit disaster appraisals into a quantifiable desire set via the B→D transition, providing explicit psychological inputs for subsequent panic arousal. The LLM serves solely as a reasoning engine for personalized PPDTS responses, avoiding the black-box instability of direct end-to-end prediction while retaining personalization and interpretability.

\subsection{Calculation of Panic Arousal Timing}
After computing desire activation and intensity, we propose a D→E state transition mechanism based on desire satisfaction evaluation, transforming multi-desire activation states into binary panic output, achieving computable triggering from intrinsic motivation to panic emotion. The core of this process quantifies the activated desire set and its intensity as a probabilistic judgment of panic occurrence at each time step.

Specifically, panic arousal probability is computed as a path-weighted product along the D→E edge of the BDEI graph, aggregating two activation conditions: (1) the proportion of activated desires $N_d$ to all 3 desire types, reflecting motivation breadth; and (2) the average activated desire intensity normalized by the PPDTS score range (1–5), representing urgency of psychological demands. Thus, $P_{panic}$ is defined as:

\begin{equation}
    P_{\text{panic}} = \frac{N_d}{3} \cdot \frac{\bar{I}_d}{5},
\end{equation}
where $\bar{I}_d$ is the average activated desire intensity, obtained by weighted averaging of individual desire intensities. 

After obtaining $P_{\text{panic}}$, binary emotion is sampled via $r\sim U(0,1)$: panic is triggered ($E_{\text{panic}} = 1$) if $r < P_{\text{panic}}$, otherwise calm ($E_{\text{panic}} = 0$). Particularly, when $N_d = 0$ (no desires activated), $P_{\text{panic}}$ is directly set to 0, maintaining a calm state (formalized in Eq. ~\ref{eq.6}). This transforms the desire–reality gap into a probabilistic emotion trigger rather than a rigid rule. The stochastic element accounts for uncertainties not captured by the cognitive model, since no model can exhaustively represent all factors influencing individual emotion.

\begin{equation}
    E = 
\begin{cases} 
1, & \text{if } r < P_{\text{panic}} \text{ and } N_d > 0, \\
0, & \text{otherwise}
\end{cases},
\label{eq.6}
\end{equation}




In summary, PCP couples multi-domain inputs via a PSD model and decomposes implicit emotion arousal into traceable state transitions along the BDEI pathway. The cognitive graph governs all transitions, while the LLM serves only for Belief-to-Desire parameter inference. This design transforms multi-domain data into dynamic inference from external stimuli to internal emotion, enabling precise panic arousal timing prediction and providing a robust foundation for proactive extreme emotion prevention in emergency social simulations.
\section{Experiments}
\subsection{Experimental Setup}
\subsubsection{Dataset Construction}


Our experiments use Hurricane Sandy (2012) as the event context. We adopt the COPE dataset \cite{liu2025pychoagent}, a fine-grained panic emotion benchmark built through a human-LLM collaborative annotation pipeline on a public hurricane dataset \cite{kryvasheyeu2015performance}. COPE partitions each user's timeline into a pre-disaster phase (242,363 tweets before landfall) and a post-disaster phase (1,142,626 tweets after landfall). Pre-disaster data is used solely for extracting individual features as model inputs; post-disaster data is used solely for annotating binary panic labels as prediction targets. This temporal partition ensures that all model inputs precede the prediction window, preventing information leakage. COPE was constructed by a human–LLM collaborative procedure with three stages: (i) initial labeling by ChatGPT-4o under an appraisal-grounded prompt framework that produces per-tweet relevance and panic judgments; (ii) two rounds of independent human review that refined the labels against appraisal-grounded annotation rules, with disagreements adjudicated by a third annotator; and (iii) propagation of the refined labels to the remaining posts by a fine-tuned discriminator, with both training and validation splits restricted to the human-verified subset. Inter-annotator agreement across the human review rounds was substantial (simple agreement = 0.9797, Cohen's $\kappa$ = 0.7059). The annotation model and PCP's inference model belong to different model families, so the labels cannot inherit model-specific biases from the inference pipeline. Construction details, including prompt design and adjudication protocols, are provided in \cite{liu2025pychoagent}, App. A.2. The four-domain inputs are:

\begin{itemize}
    \item \textbf{Physical domain:} Hurricane spatiotemporal features, from zoom.earth, social platforms, and official media, including 115 records of grade, wind speed, pressure, coordinates, and events descriptions. The event timeline is discretized into 3-hour time steps anchored at landfall.
    \item \textbf{Social domain:} User follow relationships reconstructed from the Sandy dataset, containing tens of millions of connections. The social panic ratio at time $t$ is computed from previously predicted agent states, not ground-truth labels, preserving temporal causality.
    \item \textbf{Cognitive domain:} User profiles mined from pre-disaster posts, including Big Five personality traits, historical emotion trends (neutral/positive/negative sentiment), tone features, topic preferences, and posting frequency. These features are fixed before the prediction window. The processed cognitive dataset contains 9,065 users with complete profiles.
    \item \textbf{Information domain:} Pre-disaster posts with text, timestamps, user IDs, emotion annotations, geolocation. Post-disaster posts are excluded from all model inputs.
\end{itemize}

The COPE dataset provides per-step binary panic labels derived from post-disaster tweets, which we convert into three classes using the hurricane landfall time (October 30, 2012) as a temporal anchor: users who never exhibit a panic label are classified as non-panic; those whose first panic label coincides with the landfall time step are immediate-panic; those whose first panic label appears in later time steps are delayed-panic. This conversion uses only label timestamps, not feature inputs. Table\ref{tab:dataset_stats} presents the dataset statistics.

\begin{table}[t!]
\caption{Dataset Statistics Overview}
\label{tab:dataset_stats}
\centering
\small
\begin{tabularx}{\columnwidth}{>{\centering\arraybackslash}X>{\centering\arraybackslash}X}
\toprule
\textbf{Statistical Dimension} & \textbf{Value} \\
\midrule
Number of users          & 9,065             \\
Total tweets             & 1,384,989         \\
Pre-disaster tweets      & 242,363           \\
Post-disaster tweets     & 1,142,626         \\
Physical domain records  & 115               \\
Panic users (Immediate)  & 2,164 (23.87\%)   \\
Panic users (Delayed)    & 1,076 (11.87\%)   \\
Non-panic users          & 5,825 (64.26\%)   \\
\bottomrule
\end{tabularx}
\end{table}

\subsubsection{Baseline Models}
We compare PCP with four categories of baselines. All use the same 9,065-user dataset, 115 time steps, and evaluation protocol.

\begin{itemize}
    \item \textbf{Traditional dynamics models}
\end{itemize}

SLIRS\cite{wang2016computational}: Epidemic contagion analogy with Susceptible-Latent-Infected-Recovered states. We use original parameters ($\mu$=0.01, $\eta$=0.05, $\rho$=2, $\omega$=0.9) with 2\% seed users.

BASS\cite{bass1969new}: Product diffusion driven by imitation and external innovation coefficients ($p$, $q$) calibrated via grid search to match the real peak time.

Voter\cite{muslim2024mass}: An opinion dynamics model where panic is an opinion state updated by stochastic neighbor interaction, with $q$=8, $p$=0.03, $\varepsilon$=0.05.

\begin{itemize}
    \item \textbf{Agent-Based Simulation Models}
\end{itemize}

ABM-HybridSpaces\cite{yin2024agent}: Mesa-based agents with demographics, locations, and three-layer networks. 3\% seed users with rule-based hurricane intensity response.

Crowd\cite{rende2025crowd}: An ABM framework for social network topology, mapping agent attributes to node parameters for lightweight modeling and sharing the same 9,065-node graph with PCP.

SOAR\cite{wu2018modelling}: Cognitive architecture with working memory, rules, and chunking learning. Train on first 70\% time steps, test on last 30\%. Rule libraries learned from training.

\begin{itemize}
    \item \textbf{End-to-End Prediction Models}
\end{itemize}

E-USIM \cite{naskar2020predicting}: Uses RNN and GRU with historical and current neighbor/personal emotion influence to predict next emotion state. 80/20 random user split with time-series cross-validation.

GODEN\cite{wang2024information}: User-cascade bipartite graph via GNN, Neural ODE, and Transformer for next-influencer prediction. Same social graph and user features as PCP.

\begin{itemize}
    \item \textbf{Reasoning-Enhanced Models}
\end{itemize}

ECR-Chain\cite{huang2024ecr}: LLM causal reasoning guided by cognitive appraisal theory. Uses Qwen 2.5-72B with temperature=0; input features restricted to tweet text and sentiment. 

LAIDSim\cite{hu2024llm}: LLM-enhanced independent cascade diffusion with LLM-generated profiles, BERT-encoded influence, and agent-profile-based semantic mutation. Default parameters ($\alpha$=0.5, $k$=2 seed nodes) from the original paper.

Simple Prompt: Uses template prompts for direct LLM emotion classification, without Chain-of-Thought reasoning.

CPM-RL\cite{zhang2024modeling}: Reinforcement learning with cognitive appraisal theory for emotion prediction. Q-learning agents with big-five-mapped hyperparameters and SVM classifier (linear kernel, $C$=0.014); trained on theoretical MDP distributions.

\subsubsection{Evaluation Metrics}
To comprehensively evaluate the PCP framework on panic arousal prediction, we construct an evaluation scheme from two dimensions: macro-level propagation trends and micro-level user classification.

\begin{itemize}
    \item \textbf{Macro-level} 
\end{itemize}

Macro-level metrics measure the model's ability to capture overall panic propagation dynamics, including peak scale and temporal position of the panic curve.

\textbf{Peak Count:} The maximum number of panicking agents during the simulation, reflecting propagation intensity. Smaller deviation from ground truth indicates better performance.

\textbf{Peak Time Step:} The time step at which the panic curve reaches its peak, capturing temporal characteristics and propagation speed. Closer to ground truth is better.

\textbf{Peak Count Error:} Measures prediction accuracy on peak panic scale, expressed as relative error (\%); lower is better, where $I(t)$ denotes the number of panicking agents at time $t$.
\begin{equation}
    \frac{\left| \max(I_{\text{sim}}(t)) - \max(I_{\text{real}}(t)) \right|}{\max(I_{\text{real}}(t))} \times 100\%,
\end{equation}

\textbf{Peak Time Error:} Measures prediction deviation on peak panic timing, capturing the framework's ability to capture emotion arousal dynamics. Lower is better.

\begin{equation}
    \left|\operatorname{argmax}(I_{\text{sim}}(t)) - \operatorname{argmax}(I_{\text{real}}(t))\right|,
\end{equation}

\begin{itemize}
    \item \textbf{Micro-level}
\end{itemize}

Micro-level metrics evaluate the model's ability to classify individual panic types.

\textbf{Accuracy:} Proportion of correctly classified samples across all users, i.e., the overall correct prediction rate across three panic types.

\textbf{Macro F1:} Equal-weighted average of per-class F1 scores, mitigating class-imbalance bias and emphasizing minority class (delayed panic) recognition.

\textbf{Precision (per class):} Proportion of correctly classified samples among those predicted as a given class.

\textbf{Recall (per class):} Proportion of correctly classified samples among all ground-truth samples of a given class.

\textbf{F1 (per class):} Harmonic mean of the precision and
recall of a given class.

\subsubsection{Implementation Details}

All experiments were conducted on a cloud computing environment equipped with two NVIDIA RTX4090 GPUs. LLM inference employed the Qwen2.5-72B-Instruct model, deployed via a GPU cloud function platform that provides an OpenAI-compatible API. During inference, temperature was set to 0.3 and max\_tokens to 1500. The complete list of dataset- and implementation-specific parameters, including the category-to-loss mapping, the post-dissipation loss decay schedule, the spatial and social distance handling after storm dissipation, and the LLM deployment configuration, is provided in Appendix~\ref{subsec:params}. The psychological safety distance perception threshold ${{\theta}_{risk}}$ was set to 0.08, derived by substituting the profile of an average user at the boundary of risk awareness (3-day landfall window, 900 km from storm center, 60\% of followed users panicked, Category 1 hurricane) into the PSD formula (~\ref{eq.4}). The desire arousal threshold was set to 3.5, defined on the 1–5 factor scale to which raw 1–4 PPDTS responses are linearly remapped. This value is above the neutral midpoint (3.0), ensuring that only clearly elevated appraisal states trigger desire activation rather than ambiguous or marginal states.


\begin{figure}[t!]
\centering
\includegraphics[width=\columnwidth]{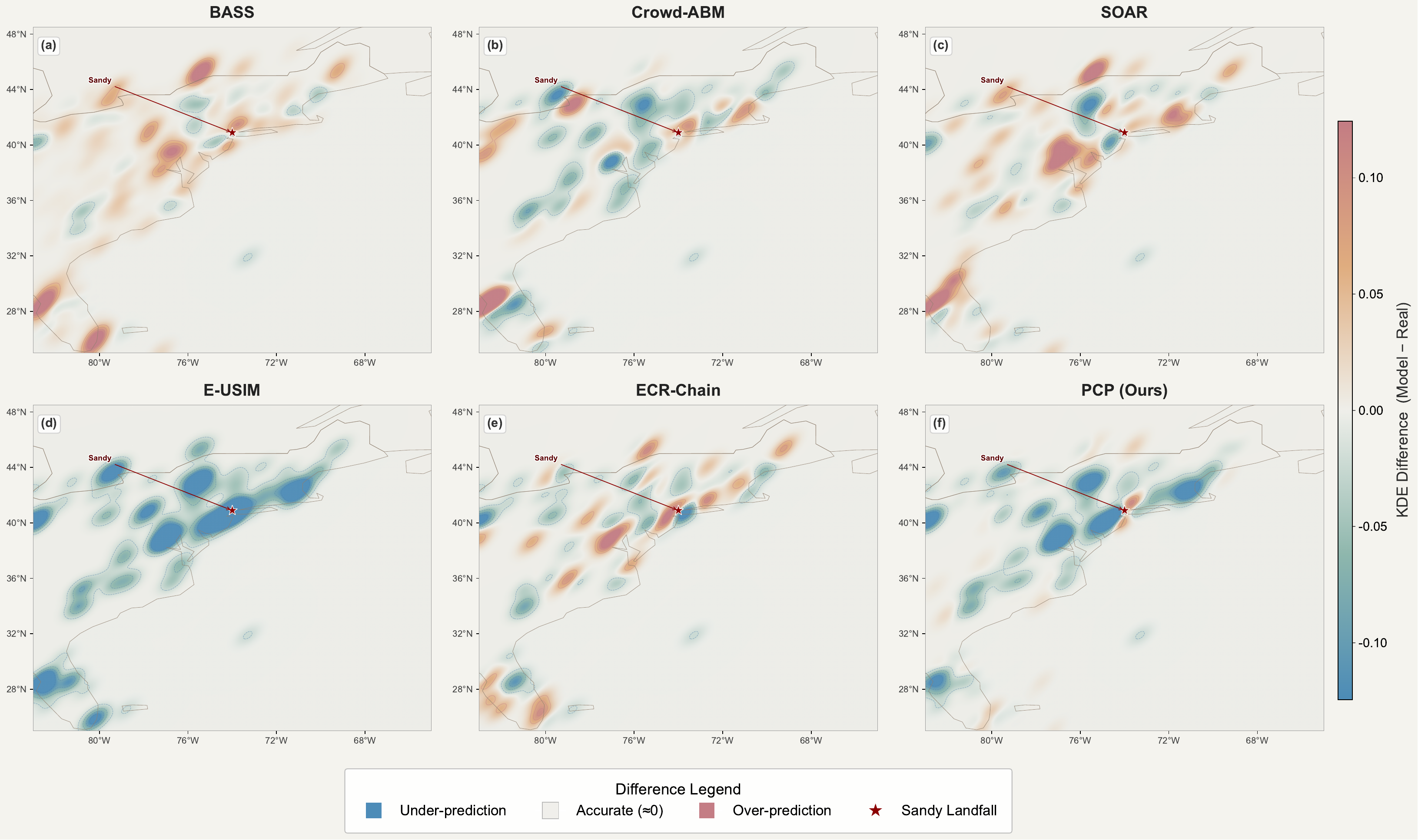}
\caption{\textbf{Spatial prediction error of peak panic arousal at Step 61 during Hurricane Sandy.} Each panel shows kernel
density estimation (KDE) difference (simulated vs. real panic density). Blue: under-prediction; Red: over-prediction; White: accurate match. All KDE heatmaps normalized to [0,1].}
\label{fig:spatial_peak}
\end{figure}

\begin{table*}[t!]
\caption{Comparison of macro-level propagation dynamics across all baseline models}
\label{tab:macro-propagation}
\centering
\footnotesize
\begin{tabularx}{\textwidth}{
    >{\hsize=1.0\hsize\centering\arraybackslash}X
    >{\hsize=1.0\hsize\centering\arraybackslash}X
    >{\hsize=0.5\hsize\centering\arraybackslash}X
    >{\hsize=0.5\hsize\centering\arraybackslash}X
    >{\hsize=0.5\hsize\centering\arraybackslash}X
    >{\hsize=0.5\hsize\centering\arraybackslash}X
}
\toprule
\textbf{Category} & \textbf{Model} & \textbf{Peak Count} & \textbf{Peak Time Step} & \textbf{Peak Error ($\downarrow$)} & \textbf{Time Error ($\downarrow$)} \\
\midrule
\multirow{3}{*}{Traditional Dynamical} & SLIRS    & 186  & 1   & 72.03\% & 60  \\
& BASS     & 231  & 45  & 65.26\% & 16  \\
& Voter    & 90   & 0   & 86.47\% & 61  \\
\midrule
\multirow{3}{*}{Agent-Based Simulation} & ABM-HybridSpaces & 271 & 0   & 59.25\% & 61  \\
& Crowd            & 308 & 59  & 53.68\% & \underline{2} \\
& SOAR             & 939 & 58  & 41.20\% & 3   \\
\midrule
\multirow{2}{*}{End-to-End Prediction}  & E-USIM  & 180  & 68  & 72.93\% & 7   \\
& GODEN   & 19   & 0   & 97.14\% & 61  \\
\midrule
\multirow{4}{*}{Reasoning-Enhanced}     & ECR-Chain     & 790  & 61  & \underline{18.80\%} & \textbf{0} \\
& LAIDSim       & 6985 & 6   & 950.38\%& 55  \\
& Simple Prompt & 3187 & 57  & 379.25\%& 4   \\
& CPM-RL        & 524  & 0   & 21.20\% & 61  \\
\midrule
\multicolumn{2}{c}{\textbf{PCP (Ours)}} & \textbf{712} & \textbf{63} & \textbf{7.07\%} & \underline{2} \\
\bottomrule
\multicolumn{6}{l}{$^{\mathrm{a}}$The ground-truth panic peak occurs at time step 61 with 665 users.} \\
\multicolumn{6}{l}{$^{\mathrm{b}}$Bold indicates the best performance in each column. Underlined indicates the second-best performance.}
\end{tabularx}
\end{table*}

\begin{table*}[t!]
\caption{Comparison of micro-level panic classification performance across all baseline models}
\label{tab:user-classification}
\centering
\resizebox{\textwidth}{!}{%
\begin{tabular}{cc cccc ccccc cccc}
\toprule
 & & \multicolumn{2}{c}{\textbf{Overall}} & \multicolumn{3}{c}{\textbf{Non-Panic}} & \multicolumn{3}{c}{\textbf{Immediate Panic}} & \multicolumn{3}{c}{\textbf{Delayed Panic}} \\
\cmidrule(lr){3-4} \cmidrule(lr){5-7} \cmidrule(lr){8-10} \cmidrule(lr){11-13}
\textbf{Category} & \textbf{Model} & \textbf{Acc.($\uparrow$)} & \textbf{Macro F1($\uparrow$)} & \textbf{Pre.($\uparrow$)} & \textbf{Rec.($\uparrow$)} & \textbf{F1($\uparrow$)} & \textbf{Pre.($\uparrow$)} & \textbf{Rec.($\uparrow$)} & \textbf{F1($\uparrow$)} & \textbf{Pre.($\uparrow$)} & \textbf{Rec.($\uparrow$)} & \textbf{F1($\uparrow$)} \\
\midrule
\multirow{3}{*}{Traditional Dynamical Models} & SLIRS    & 0.6313 & 0.2709 & 0.6418 & \underline{0.9746} & 0.7740 & 0.2091 & 0.0213 & 0.0396 & - & - & - \\
& BASS     & 0.2276 & 0.1618 & \textbf{0.8889} & 0.0014 & 0.0027 & 0.2381 & \underline{0.9025} & 0.3797 & 0.1194 & \underline{0.0948} & 0.1057 \\
& Voter    & 0.6057 & 0.2924 & 0.6435 & 0.9112 & 0.7544 & 0.2240 & 0.0846 & 0.1228 & - & - & - \\
\midrule
\multirow{3}{*}{Agent-Based Simulation Models} & ABM-HybridSpaces & 0.6449 & 0.2907 & 0.6667 & 0.9488 & 0.7831 & 0.2592 & 0.0518 & 0.0864 & 0.0667 & 0.0014 & 0.0027 \\
& Crowd           & 0.5732 & 0.3196 & 0.6439 & 0.8220 & 0.7221 & 0.2552 & 0.1821 & 0.2125 & 0.1647 & 0.0130 & 0.0241 \\
& SOAR            & 0.2424 & 0.1473 & 0.7248 & 0.0136 & 0.0266 & 0.2420 & \textbf{0.9700} & 0.3874 & 0.0671 & 0.0177 & 0.0280 \\
\midrule
\multirow{2}{*}{End-to-End Prediction Models} & E-USIM  & \underline{0.6498} & 0.3483 & 0.6798 & 0.9373 & \underline{0.7880} & 0.3833 & 0.1635 & 0.2292 & \underline{0.5000} & 0.0142 & 0.0276 \\
& GODEN   & 0.6434 & 0.2643 & 0.6435 & \textbf{0.9993} & 0.7829 & \textbf{0.5789} & 0.0051 & 0.0101 & - & - & - \\
\midrule
\multirow{4}{*}{Reasoning-Enhanced Models}    & ECR-Chain    & 0.5045 & \underline{0.4419} & 0.8432 & 0.5298 & 0.6507 & 0.3848 & 0.4492 & 0.4145 & 0.1789 & \textbf{0.4786} & \textbf{0.2604} \\
& LAIDSim      & 0.3758 & 0.2665 & 0.7644 & 0.2730 & 0.4023 & 0.2601 & 0.8396 & 0.3972 & - & - & - \\
& Simple Prompt& 0.4802 & 0.3403 & 0.7642 & 0.4702 & 0.5822 & 0.2976 & 0.7421 & \underline{0.4248} & 0.0952 & 0.0074 & 0.0138 \\
& CPM-RL       & 0.6142 & 0.2796 & 0.6396 & 0.9379 & 0.7605 & 0.2004 & 0.0485 & 0.0781 & - & - & - \\
\midrule
\multicolumn{2}{c}{\textbf{PCP (Ours)}} & \textbf{0.7566} & \textbf{0.5441} & \underline{0.8723} & 0.8736 & \textbf{0.8730} & \underline{0.5398} & 0.7860 & \textbf{0.6401} & \textbf{0.8625} & 0.0641 & \underline{0.1194} \\
\bottomrule
\multicolumn{13}{l}{$^{\mathrm{a}}$Bold indicates the best performance in each column. Underlined indicates the second-best performance.}\\
\multicolumn{13}{l}{$^{\mathrm{b}}$A dash (-) indicates that the model assigned no instances to the delayed panic class, resulting in undefined precision, recall, and F1.}
\end{tabular}
}
\end{table*}

\subsection{Comparison with SOTA Baselines}
We evaluate PCP on Hurricane Sandy against baselines. PCP reduces peak count error to 7.07\% and peak time deviation to 2 steps (Table~\ref{tab:macro-propagation}). Traditional dynamical models assign identical transition rules to all individuals, so a person who panics late from a distant trigger is treated the same as one who panics immediately, yielding peak errors of 65-86\%. Agent-based simulations allow individual differences but depend on handcrafted rules, and even SOAR's best 41.20\% cannot capture emergent social media behaviors that violate these rules. Reasoning-enhanced methods let LLMs generate steps without structural constraints, producing exaggerated outputs with peak errors of 379-950\%. Figure~\ref{fig:spatial_peak} confirms this spatially. PCP's KDE difference at Step 61 is near-zero across most regions, while BASS and ECR-Chain over-predict along the East Coast and E-USIM under-predicts. This fidelity arises from PSD's four-domain fusion, which lets each agent evaluate psychological safety distance so that a critical mass crossing the threshold naturally produces the observed peak.

PCP achieves 75.66\% accuracy and Macro F1 0.5441, leading all baselines by 10.68pp (Table~\ref{tab:user-classification}). Per-class F1 is 0.8730 (non-panic), 0.6401 (immediate), and 0.1194 (delayed). Traditional models assume irreversible or consensus-driven transitions, predicting nearly all users as non-panic (e.g., SLIRS recall 0.02) or as panicked (e.g., BASS recall 0.90). Agent-based simulations similarly default to extremes (e.g., SOAR recall 0.97 versus ABM-HybridSpaces recall 0.05) without cognitive appraisal to differentiate individual threat evaluation. End-to-end models learn from past observations, so delayed panic cases that emerge after the initial hazard subsides appear as non-panic, yielding recall below 1.5\% for E-USIM and GODEN. Among reasoning-enhanced methods, ECR-Chain over-assigns delayed labels with precision 0.1789 versus recall 0.4786. PCP avoids these extremes by explicitly modeling the appraisal-to-arousal transition, achieving balanced recall (0.87 for Non-Panic, 0.79 for Immediate Panic).


As difficulty increases from non-panic to immediate to delayed panic, PCP's F1 score drops from 0.8730 to 0.6401 to 0.1194. On delayed panic, PCP achieves the highest precision (0.8625) across all models, providing reliable preliminary identification rather than comprehensive coverage. This challenge is shared across all baselines, none of which model post-subsidence arousal triggers. Traditional models use fixed equations with no mechanism for external triggers; end-to-end models learn from past observations, but these delayed factors lie in the future. SOAR's rule base covers only the first 70\% of time steps; CPM-RL evaluates appraisal independently per step without retaining prior evaluations, so intensity cannot accumulate. ECR-Chain over-assigns delayed labels (precision 0.18 vs. recall 0.48). No baseline integrates hazard context, cognitive reappraisal, delayed factors, and peer influence with explicit cognitive pathway modeling.

\begin{table}[t!]
\caption{Delayed-panic first-tweet thematic composition ($n=1{,}076$). ``Canonical exemplars" are the most frequent surface forms per category.}
\label{tab:delayed-composition}
\setlength{\tabcolsep}{6pt} 
\renewcommand{\arraystretch}{1.15}
\footnotesize
\begin{tabular}{@{}m{0.15\linewidth} m{0.48\linewidth} r r@{}}
\toprule
\textbf{Category} & \textbf{Canonical exemplars}
 & \textbf{$n$ users} & \textbf{share} \\
\midrule
\textbf{Secondary}
 & power outage, evacuation, damage
 & \textbf{539} & \textbf{50.09\%} \\
Primary
 & Hurricane Sandy, landfall, the storm
 & 421 & 39.13\% \\
Other
 & ---
 & 113 & 10.50\% \\
Social
 & my family, praying, stay safe
 & 3   & 0.28\%  \\
\midrule
\textbf{Total}
 & ---
 & \textbf{1076} & \textbf{100.00\%} \\
\bottomrule
\end{tabular}
\end{table}

To characterize the conservative recall (0.0641) on delayed panic, we conducted a lexicon-based composition analysis of all 1,076 delayed-panic users' first panic expressions (Table~\ref{tab:delayed-composition}), using the four-category lexicon and priority rules reproduced in Appendix~\ref{subsec:lexicon}. 50.09\% reference secondary hazards that emerge after landfall, including flooding, power outages, and resource scarcity; 39.13\% reference the primary storm; 0.28\% reference social/peer signals; and 10.50\% are other. Within PCP, the social and informational dimensions of PSD can sustain perceived risk after the primary hazard subsides, since the proportion of panicking users in an agent's attention list continues to drive perceived risk; this explains why the delayed cases PCP does identify are predominantly socially triggered. The unidentified remainder is driven by secondary hazards that are absent from PCP's input stream by construction, since all features are fixed before the prediction window and the physical domain encodes only the primary-hazard trajectory. Predicting such cases would require inputs that do not yet exist at prediction time. This is as infeasible as forecasting rainfall with no signal of future clouds.

\begin{table*}[t!]
\caption{Ablation study results: macro-level metrics}
\label{tab:ablation-macro}
\centering
\footnotesize
\begin{tabularx}{\textwidth}{>{\centering\arraybackslash}X>{\centering\arraybackslash}X>{\centering\arraybackslash}X>{\centering\arraybackslash}X>{\centering\arraybackslash}X>{\centering\arraybackslash}X}
\toprule
\textbf{Model} & \textbf{Peak Panic Count} & \textbf{Peak Time Step} & \textbf{Peak Error ($\downarrow$)} & \textbf{Time Error ($\downarrow$)} & \textbf{Time Cost} \\
\midrule
w/o BDEI Path & 3187 & 57 & 379.25\% & 4 & 115h56min \\
w/o PSD & 184 & 63 & 72.33\% & 2 & 18h30min \\
w/o LLM & 2685 & 56 & 303.76\% & 5 & 1h30min \\
Full & 712 & 63 & 7.07\% & 2 & 14h27min \\
\bottomrule
\end{tabularx}
\end{table*}

\begin{table*}[t!]
\caption{Ablation study results: micro-level metrics}
\label{tab:ablation-micro}
\centering
\footnotesize
\begin{tabularx}{\textwidth}{
    >{\hsize=1.2\hsize\centering\arraybackslash}X
    >{\hsize=1.2\hsize\centering\arraybackslash}X
    >{\hsize=1.4\hsize\centering\arraybackslash}X   
    >{\hsize=1.0\hsize\centering\arraybackslash}X
    >{\hsize=0.9\hsize\centering\arraybackslash}X
    >{\hsize=0.9\hsize\centering\arraybackslash}X   
    >{\hsize=0.9\hsize\centering\arraybackslash}X
    >{\hsize=0.9\hsize\centering\arraybackslash}X
    >{\hsize=0.9\hsize\centering\arraybackslash}X   
    >{\hsize=0.9\hsize\centering\arraybackslash}X
    >{\hsize=0.9\hsize\centering\arraybackslash}X
    >{\hsize=0.9\hsize\centering\arraybackslash}X   
}
\toprule
\multirow{2}{*}{\textbf{Model}} & \multicolumn{2}{c}{\textbf{Overall}} & \multicolumn{3}{c}{\textbf{Non-Panic}} & \multicolumn{3}{c}{\textbf{Immediate Panic}} & \multicolumn{3}{c}{\textbf{Delayed Panic}} \\
\cmidrule(lr){2-3} \cmidrule(lr){4-6} \cmidrule(lr){7-9} \cmidrule(lr){10-12}
 & \textbf{Acc. ($\uparrow$)} & \textbf{F1 ($\uparrow$)} & \textbf{Pre. ($\uparrow$)} & \textbf{Rec. ($\uparrow$)} & \textbf{F1 ($\uparrow$)} & \textbf{Pre. ($\uparrow$)} & \textbf{Rec. ($\uparrow$)} & \textbf{F1 ($\uparrow$)} & \textbf{Pre. ($\uparrow$)} & \textbf{Rec. ($\uparrow$)} & \textbf{F1 ($\uparrow$)} \\
\midrule
w/o BDEI & 0.4802 & 0.3403 & 0.7642 & 0.4702 & 0.5822 & 0.2976 & 0.7421 & 0.4248 & 0.0952 & 0.0074 & 0.0138 \\
w/o PSD & 0.6138 & 0.3611 & 0.6693 & 0.8539 & 0.7504 & 0.3715 & 0.2666 & 0.3105 & 0.1625 & 0.0121 & 0.0225 \\
w/o LLM & 0.2293 & 0.1816 & 0.7167 & 0.0074 & 0.0146 & 0.2579 & 0.8489 & 0.3956 & 0.1057 & 0.1849 & 0.1345 \\
Full & 0.7566 & 0.5441 & 0.8723 & 0.8736 & 0.8730 & 0.5398 & 0.7860 & 0.6401 & 0.8625 & 0.0641 & 0.1194 \\
\bottomrule
\end{tabularx}
\end{table*}

\subsection{Ablation Studies}
To verify the contribution of each core component in the PCP framework, we design a three-level ablation study: (1) w/o BDEI Path, removes the cognitive pathway structure and uses direct end-to-end panic prediction via LLM with user profiles and disaster context; (2) w/o PSD, removes the PSD model and replaces it with a physical distance threshold for risk perception; (3) w/o LLM, removes LLM-assisted reasoning and replaces PPDTS factor scoring with rule-based mapping from raw Big Five trait scores and contextual features. Results are shown in Tables ~\ref{tab:ablation-macro} and ~\ref{tab:ablation-micro}.

Results at the macro level show the full model achieves a peak error of 7.07\% and a timing deviation of 2 steps. Removing either the BDEI pathway (peak error 379.25\%) or the LLM (303.76\%) causes the simulated panic count to explode by 4-5$\times$, indicating that both structured cognitive reasoning and LLM-augmented inference are necessary to prevent uncontrolled panic propagation. By contrast, removing PSD yields a milder degradation (72.33\%), as physical distance alone provides a rough but insufficient spatial signal for risk differentiation. In terms of computational efficiency, PCP reduces inference time to 14h 27min ($\approx 8\times$ faster than the pure LLM variant), as PSD filters out low-risk users before they enter the BDEI pathway, reducing unnecessary LLM invocations while preserving multi-domain risk perception.

At the micro level, the full model achieves the best overall accuracy (0.7566) and Macro F1 (0.5441). Each ablation exposes a distinct failure mode: w/o LLM over-predicts immediate panic (recall: 0.8489, precision: 0.2579) due to rigid rule-based thresholds; w/o PSD captures non-panic well (recall: 0.8539) but misses immediate panic (recall: 0.2666), confirming physical distance alone lacks cognitive granularity; w/o BDEI degrades uniformly (Acc: 0.4802) as direct LLM prediction loses the discriminative reasoning chain. Delayed panic remains a shared challenge ($F_1<0.14$ across all variants). The full model achieves the highest precision (0.8625) with a low recall (0.0641), reflecting conservative yet reliable judgments, whereas w/o LLM trades precision for recall. This precision-recall trade-off reflects the framework's design scope: PCP prioritizes reliable identification over broad coverage on delayed panic, an important direction for future improvement.


Collectively, the ablation results validate the PCP design: BDEI ensures theoretical completeness, multi-domain PSD captures dynamic risk perception, and LLM enables sensitivity to individual heterogeneity. Their synergy drives the full model's overall superiority over all ablations, with the lowest false positive rate on delayed panic recognition.

\subsection{Robustness Analysis}
Since the BDEI scoring module relies on stochastic LLM inference, PPDTS dimension scores must remain consistent across repeated calls under identical inputs. We quantify robustness via the within-group coefficient of variation ($CV~=~\sigma/\mu$) over $M=30$ independent runs per (model, temperature, user) triplet, averaged across 10 users. Lower CV indicates higher scoring stability. As no universal CV threshold exists for LLM-based scoring, robustness is evaluated through internal comparison against both the deterministic baseline ($T=0.0$) and alternative models.

\begin{figure}[t!]
\centering
\includegraphics[width=\columnwidth]{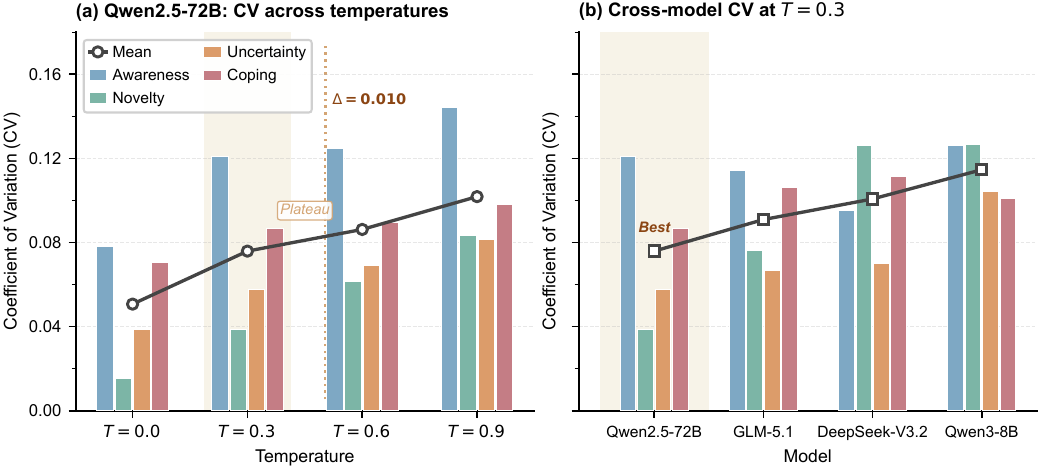}
\caption{Robustness analysis of LLM scoring consistency. (a) Per-dimension and mean CV of Qwen2.5-72B-Instruct across temperatures. Shaded: main config $T=0.3$; $\Delta=0.010$ at $T=0.6$ indicates stable performance. (b) At $T=0.3$, Qwen2.5-72B has the lowest and most uniform CV across dimensions.}
\label{fig:robustness}
\end{figure}

\textbf{Temperature sensitivity.} Figure~\ref{fig:robustness}(a) and Table~\ref{tab:qwen_temp_cv} report per-dimension and mean CV of Qwen2.5-72B-Instruct across four temperatures. At $T=0.0$, the model is deterministic (mean CV~=~0.051), but deterministic decoding produces rigid, unnuanced responses that fail to capture the cognitive variability inherent in individual-level appraisal. The critical observation is that the CV increment from $T=0.3$ to $T=0.6$ is merely $\Delta=0.010$, indicating $T=0.3$ lies at the onset of a stability plateau. This confirms $T=0.3$ offers a favorable trade-off: sufficient sampling diversity while maintaining reproducible scoring.

\textbf{Model sensitivity.} Figure~\ref{fig:robustness}(b) and Table~\ref{tab:model_cv_t03} compare CV at $T=0.3$. Qwen2.5-72B-Instruct achieves the lowest mean CV ($0.076$), outperforming GLM-5.1 ($0.091$), DeepSeek-V3.2 ($0.101$), and Qwen3-8B ($0.115$). The advantage is most pronounced in the novelty dimension (CV $=0.039$ vs. second-best $0.076$). Conversely, Qwen3-8B exhibits the highest CV in three of four dimensions, suggesting insufficient consistency. These results jointly justify selecting Qwen2.5-72B-Instruct at $T=0.3$ as the primary configuration.

\begin{table}[t!]
\caption{CV of Qwen2.5-72B across temperatures}
\label{tab:qwen_temp_cv}
\centering
\footnotesize
\setlength{\tabcolsep}{4pt}
\begin{tabularx}{\columnwidth}{
    >{\hsize=0.5\hsize\centering\arraybackslash}X   
    >{\hsize=1.0\hsize\centering\arraybackslash}X   
    >{\hsize=1.0\hsize\centering\arraybackslash}X   
    >{\hsize=1.0\hsize\centering\arraybackslash}X   
    >{\hsize=1.0\hsize\centering\arraybackslash}X   
    >{\hsize=0.5\hsize\centering\arraybackslash}X   
}
\toprule
\textbf{Temp.} & \textbf{Awareness} & \textbf{Novelty} & \textbf{Uncertainty} & \textbf{Coping} & \textbf{Mean} \\
\midrule
0.0 & 0.0779 & 0.0154 & 0.0387 & 0.0707 & 0.0507 \\
\textbf{0.3} & \textbf{0.1208} & \textbf{0.0385} & \textbf{0.0578} & \textbf{0.0867} & \textbf{0.0759} \\
0.6 & 0.1249 & 0.0613 & 0.0693 & 0.0895 & 0.0862 \\
0.9 & 0.1441 & 0.0834 & 0.0816 & 0.0982 & 0.1018 \\
\bottomrule
\multicolumn{6}{l}{$^{\mathrm{a}}$Bold row indicates the configuration used in the main experiment.}
\end{tabularx}
\end{table}

\begin{table}[t!]
\caption{Cross-model CV comparison at $T=0.3$}
\label{tab:model_cv_t03}
\centering
\small
\setlength{\tabcolsep}{3.5pt}
\resizebox{\columnwidth}{!}{%
\begin{tabular}{lccccc}
\toprule
\textbf{Model} & \textbf{Awareness} & \textbf{Novelty} & \textbf{Uncertainty} & \textbf{Coping} & \textbf{Mean} \\
\midrule
\textbf{Qwen2.5-72B} & \textbf{0.1208} & \textbf{0.0385} & \textbf{0.0578} & \textbf{0.0867} & \textbf{0.0759} \\
GLM-5.1                & 0.1145 & 0.0764 & 0.0666 & 0.1060 & 0.0909 \\
DeepSeek-V3.2          & 0.0952 & 0.1261 & 0.0702 & 0.1115 & 0.1007 \\
Qwen3-8B               & 0.1262 & 0.1267 & 0.1045 & 0.1009 & 0.1146 \\
\bottomrule
\multicolumn{6}{l}{$^{\mathrm{a}}$Bold row indicates the model selected for the main experiment.}
\end{tabular}
}
\end{table}

\begin{figure*}[t!]
  \centering
  \includegraphics[width=1.0\linewidth]{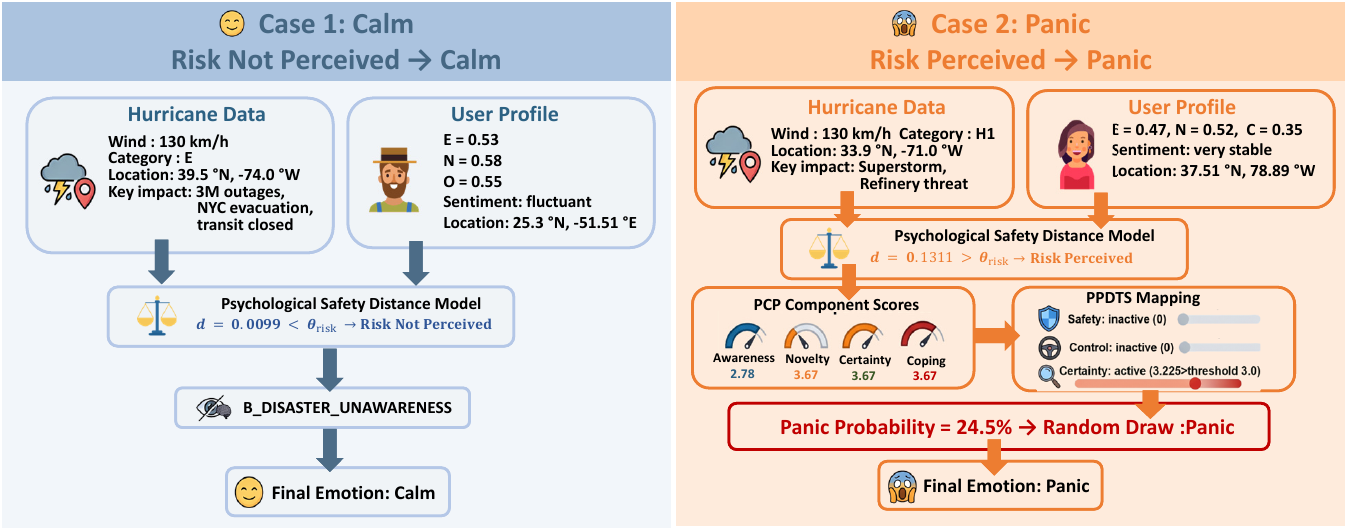}
  \caption{PCP framework predicts counterfactual panic in a lower-neuroticism user (Case 2) by accounting for situational uncertainty and a spatially shielded higher-neuroticism user (Case 1).}
  \label{fig:case_study}
\end{figure*}

\subsection{Case Study}
To validate PCP's interpretability, we analyze two counterfactual cases as illustrated in Fig.~\ref{fig:case_study}. 
Case 1: a user with higher neuroticism ($Extraversion (E) = 0.53, Neuroticism (N) = 0.58$) far from Hurricane Sandy's center. Despite susceptibility to anxiety, large physical distance reduces perceived threat, keeping psychological safety distance well below the panic threshold. The user remains calm, showing how spatial shielding decouples cognitive appraisal from dispositional vulnerability. Case 2: a user with lower neuroticism ($E=0.47, N=0.52$) in the direct impact zone. Although the stable baseline implies resilience, the convergence of hazard proximity and information escalation compressed the safety distance above threshold, overriding dispositional regulation.
Both cases show that multi-domain features interact through cognitive pathways to produce the final emotional state.

\subsection{Discussion}

The experiments yield a consistent answer. PCP faithfully reproduces macro-level collective panic emergence while achieving reliable individual recognition. This dual advantage stems from three synergistic design choices. (1) Multi-domain PSD fusion replaces the single-dimension inputs of prior work. It couples heterogeneous data streams across physical, social, cognitive, and informational domains into a unified safety distance metric. As a result, each agent evaluates threat from multiple perspectives, and the collective peak emerges naturally rather than being distorted by single-factor rules. (2) Explicit BDEI reasoning decomposes the implicit panic formation process into discrete traceable steps. These steps span from risk perception through emotion triggering to intention generation. This makes the cognitive pathway transparent and explainable, rather than a black-box mapping from inputs to labels. (3) Structured LLM constraint confines the model to parameter inference for only the Belief-to-Desire transition. This inverts the conventional role and limits hallucination propagation to single steps, while still retaining sensitivity to individual heterogeneity. We emphasize that this design mitigates rather than eliminates cross-step error flow: an incorrect desire at step $t$ may still propagate through panic probability and social panic exposure $n$ to neighbors. Per-step PSD re-evaluation and the three-desire aggregation in D$\to$E curb such cascading, as evidenced by the ablation gap between w/o BDEI (379\% peak error) and Full PCP (7.07\%). These three choices are not independent. PSD provides the multi-domain input that BDEI requires, BDEI supplies the pathway structure that constrains the LLM, and the LLM injects the individual variability that rigid cognitive rules alone cannot capture. Their synergy drives the full model's advantage over all baselines.

The same experimental design also defines the scope of the conclusions. The main limitations are as follows. First, Our composition analysis indicates that a majority of delayed-panic cases are triggered by secondary hazards emerging after the primary event subsides. PCP focuses on primary events without fully modeling secondary event interactions on panic arousal, which contributes to the conservative delayed panic recall across all models. Second, PCP has been validated only on Hurricane Sandy. Despite domain-general theoretical grounding, its reliance on these theories and PPDTS leaves cross-cultural and cross-disaster generalization unevaluated. Furthermore, the English-language Twitter data may introduce language, platform, and geographic biases by under representing non-English populations, demographics outside the platform's user base, and regions with uneven network coverage. Third, PSD discount weights and desire arousal thresholds are determined via domain knowledge and hyperparameter tuning, lacking adaptive optimization that may affect generalization under data shift. Finally, the Emotion-to-Intention transition uses a simplified model, abstracting emotional expression as state updates and propagation without modeling specific behaviors. 

These limitations point to four future directions. (1) Incorporating secondary event generation and coupling mechanisms to enable full-trajectory simulation of collective panic arousal, improving both macro-level emergent dynamics reconstruction and delayed panic recognition. (2) Extending PCP to diverse emergency types such as earthquakes and public health events, and exploring adaptation to other extreme emotions. (3) Optimizing the LLM-graph collaborative reasoning architecture with lightweight fine-tuning and dynamic threshold adaptation to improve efficiency while preserving interpretability. (4) Integrating social network behavioral logs to model the emotion-intention-behavior chain in BDEI pathways, achieving full-chain simulation from psychological states to social actions.

\section{Conclusion}

We ground panic arousal timing prediction in appraisal emotion theory and propose PanicCognitivePath (PCP) as a unified framework. PCP couples heterogeneous data across physical, social, cognitive, and informational domains via a psychological safety distance model. It extends BDI with an explicit Emotion node through a BDEI cognitive pathway. It confines the LLM to parameter inference for only the Belief-to-Desire transition to limit hallucination propagation to single steps. Experiments on Hurricane Sandy show that PCP improves individual prediction accuracy by 10.68\%, reduces peak count error to 7.07\%, consistently outperforming all baselines. Beyond these empirical gains, the broader implication is that panic arousal prediction can be treated as structured cognitive pathway reasoning rather than only as behavioral pattern fitting or unconstrained LLM generation. More generally, the proposed framework suggests a design principle for psychologically grounded prediction under multi-domain inputs. Fuse heterogeneous data into unified perception metrics. Decompose the target process into traceable cognitive steps. Constrain the LLM to well-defined subtasks within an explicit pathway rather than placing it at the center of decision-making. This perspective connects psychological theory with computational modeling and may be useful for other social simulation problems in which the target process has theoretical structure, multi-domain inputs must be integrated, and generative models need structural guardrails to remain reliable.

\section*{ACKNOWLEDGMENTS}
We thank all collaborators for their invaluable contributions to this project. We thank Prof. Bin Chen's group from the University of Electronic Science and Technology of China for their support of this work. We thank Assoc. Prof. Lingnan He from Sun Yat-sen University and Prof. Kaisheng Lai from Jinan University for their professional guidance in psychology. We disclose that the example social media posts illustrated in the Information Domain of Figure~\ref{fig:pcp_framework_in_hurricane_disaster} were generated by the Nano Banana2 model and do not represent real user content.

\bibliographystyle{ieeetr}
\bibliography{custom}

@article{sheikhi2025looting,
  title={Looting and antisocial behavior after disasters: a systematic review},
  author={Sheikhi, Rahim Ali and Javanbakhtian, Raheleh and Heidari, Mohammad},
  journal={BMC Public Health},
  volume={25},
  number={1},
  pages={309},
  year={2025},
  publisher={Springer}
}

@article{garfin2014exposure,
  title={Exposure to rapid succession disasters: a study of residents at the epicenter of the {C}hilean {B}{\'\i}o {B}{\'\i}o earthquake},
  author={Garfin, Dana Rose and Silver, Roxane Cohen and Ugalde, Francisco Javier and Linn, Heiko and Inostroza, Manuel},
  journal={J. Abnorm. Psychol.},
  volume={123},
  number={3},
  pages={545-556},
  year={2014},
  publisher={American Psychological Association}
}

@article{wu2025novel,
title={A Novel Agent-Based Approach for Dynamic Emotion Modeling in Social Networks},
author = {Xiaokun Wu and Limeng Lu and Mariagrazia Dotoli and Giancarlo Fortino and Min Chen},
journal={IEEE Trans. Cybern.},
volume={56},
number={3},
pages={1365-1378},
year={2026},
}

@article{adam2016bdi,
  title={{BDI} agents in social simulations: a survey},
  author={Adam, Carole and Gaudou, Benoit},
  journal={Knowl. Eng. Rev.},
  volume={31},
  number={3},
  pages={207-238},
  year={2016},
  publisher={Cambridge Univ. Press}
}

@inproceedings{shen2025coe,
  title={{CoE}: A clue of emotion framework for emotion recognition in conversations},
  author={Shen, Zhiyu and Pang, Yunhe and Rao, Yanghui and Yu, Jianxing},
  booktitle={Proc. 63rd Annu. Meet. Assoc. Comput. Linguist. (ACL) (Volume 1: Long Papers)},
  pages={23548-23563},
  year={2025}
}

@article{wang2016computational,
  title={Computational models and optimal control strategies for emotion contagion in the human population in emergencies},
  author={Wang, Xiaoming and Zhang, Lichen and Lin, Yaguang and Zhao, Yanxin and Hu, Xiaolin},
  journal={Knowl.-Based Syst.},
  volume={109},
  pages={35-47},
  year={2016},
  publisher={Elsevier}
}

@article{wang2025modeling,
  title={Modeling the spread of negative emotions in social networks during sudden public crisis events: Dual mechanisms of social reinforcement and individual regulation},
  author = {Wang, Tiantian and Liu, Tiezhong and Li, Congcong},
  journal={J. Tsinghua Univ. (Sci. Technol.)},
  volume={65},
  number={6},
  pages={1040-1049},
  year={2025},
  publisher={清华大学出版社}
}

@article{chu2024emotional,
  title={Emotional contagion on social media and the simulation of intervention strategies after a disaster event: A modeling study},
  author={Chu, Meijie and Song, Wentao and Zhao, Zeyu and Chen, Tianmu and Chiang, Yi-chen},
  journal={Humanit. Soc. Sci. Commun.},
  volume={11},
  number={1},
  pages={968},
  year={2024},
  publisher={Springer Science and Business Media LLC}
}

@article{zhao2024improved,
title={Improved Crowd Dynamics Analysis Considering Physical Contact Force and Panic Emotional Propagation},
author={Rongyong Zhao and Bingyu Wei and Chuanfeng Han and Ping Jia and Wenjie Zhu and Cuiling Li and Yunlong Ma},
journal={IEEE Trans. Intell. Transp. Syst.},
volume={26},
number={2},
pages={1840-1851},
year={2025},
publisher={IEEE}
}

@inproceedings{park2023generative,
author = {Park, Joon Sung and O'Brien, Joseph and Cai, Carrie Jun and Morris, Meredith Ringel and Liang, Percy and Bernstein, Michael S.},
title = {Generative Agents: Interactive Simulacra of Human Behavior},
year = {2023},
isbn = {9798400701320},
publisher = {ACM},
address = {San Francisco, CA, USA},
url = {https://doi.org/10.1145/3586183.3606763},
booktitle = {Proc. 36th Annu. ACM Symp. User Interface Softw. Technol. (UIST)},
articleno = {2},
pages={1-22},
location = {San Francisco, CA, USA},
}

@inproceedings{hu2024llm,
title={An {LLM}-enhanced Agent-based Simulation Tool for Information Propagation},
author={Yuxuan Hu and Gemju Sherpa and Lan Zhang and Weihua Li and Quan Bai and Yijun Wang and Xiaodan Wang},
booktitle={Proc. 33rd Int. Joint Conf. Artif. Intell. (IJCAI)},
year={2024},
pages={8679-8682}
}

@inproceedings{wang2025yulan,
  title={{Yulan-OneSim}: Towards the next generation of social simulator with large language models},
  author={Wang, Lei and Gao, Heyang and Bo, Xiaohe and Chen, Xu and Wen, Ji-Rong},
  booktitle={Workshop on Scaling Environments for Agents},
  year={2025}
}

@inproceedings{wu2024coke,
  title={{COKE}: A cognitive knowledge graph for machine theory of mind},
  author={Wu, Jincenzi and Chen, Zhuang and Deng, Jiawen and Sabour, Sahand and Meng, Helen and Huang, Minlie},
  booktitle={Proc. 62nd Annu. Meet. Assoc. Comput. Linguist. (ACL) (Volume 1: Long Papers)},
  pages={15984-16007},
  year={2024}
}

@inproceedings{ye2025emotiongraphrag,
  title={{EmotionGraphRAG}: Enhancing {LLM}-Based Emotion Analysis through Graph Retrieval-Augmented Generation},
  author={Ye, Fangzheng},
  booktitle={Proc. 7th Int. Conf. Next Gener. Data-driven Netw. (NGDN)},
  pages={227-230},
  year={2025},
  organization={IEEE}
}

@book{ortony2022cognitive,
  title={The Cognitive Structure of Emotions},
  author={Ortony, Andrew and Clore, Gerald L and Collins, Allan},
  edition={2nd},
  year={2022},
  publisher={Cambridge Univ. Press},
  address={Cambridge, U.K.}, 

}

@article{wang2024logic,
  title={On the logic of agent’s emotions},
  author={Wang, Yuanyi and Liu, Zhen and Liu, Tingting and Samsonovich, Alexei V and Klimov, Valentin V},
  journal={Cogn. Syst. Res.},
  volume={88},
  pages={101281},
  year={2024},
  publisher={Elsevier}
}

@inproceedings{rao1995bdi,
  author = {Rao, Anand S and George, Michael P},
  booktitle = {Proc. 1st Int. Conf. Multi-Agent Syst. (ICMAS)},
  pages = {312-319},
  title = {{BDI} agents: From theory to practice},
  url = {http://www.agent.ai/doc/upload/200302/rao95.pdf},
  year = 1995
}

@inproceedings{de2020bdi,
  language = {English},
  copyright = {Compilation and indexing terms, Copyright 2026 Elsevier Inc.},
  copyright = {Compendex},
  journal = {IJCAI International Joint Conference on Artificial Intelligence},
  title={{BDI} agent architectures: A survey},
  author={De Silva, Lavindra and Meneguzzi, Felipe and Logan, Brian},
  booktitle={Proc. 29th Int. Joint Conf. Artif. Intell. (IJCAI)},
  year = {2020},
  pages = {4914-4921},
  issn = {10450823},
  address = {Yokohama, Japan},
}

@article{ong2019computational,
  title={Computational models of emotion inference in theory of mind: A review and roadmap},
  author={Ong, Desmond C and Zaki, Jamil and Goodman, Noah D},
  journal={Top. Cogn. Sci.},
  volume={11},
  number={2},
  pages={338-357},
  year={2019},
  publisher={Wiley Online Library}
}

@book{hatfield1993emotional,
address={Cambridge, U.K.}, 
series={Studies in Emotion and Social Interaction}, 
title={Emotional Contagion}, 
publisher={Cambridge Univ. Press}, 
author={Hatfield, Elaine and Cacioppo, John T. and Rapson, Richard L.}, 
year={1993}, 
collection={Studies in Emotion and Social Interaction}
}

@article{mclennan2020conceptualising,
  title={Conceptualising and measuring psychological preparedness for disaster: The Psychological Preparedness for Disaster Threat Scale},
  author={McLennan, Jim and Marques, Mathew D and Every, Danielle},
  journal={Nat. Hazards},
  volume={101},
  number={1},
  pages={297-307},
  year={2020},
  publisher={Springer}
}

@article{bullogh1912psychical,
author = {Bullough, EDWARD},
title = {`{P}SYCHICAL DISTANCE' AS A FACTOR IN ART AND AN AESTHETIC PRINCIPLE},
journal = {Br. J. Psychol.},
volume = {5},
number = {2},
pages = {87-118},
doi = {https://doi.org/10.1111/j.2044-8295.1912.tb00057.x},
url = {https://bpspsychub.onlinelibrary.wiley.com/doi/abs/10.1111/j.2044-8295.1912.tb00057.x},
eprint = {https://bpspsychub.onlinelibrary.wiley.com/doi/pdf/10.1111/j.2044-8295.1912.tb00057.x},
year = {1912}
}

@article{trope2010construal,
  title={Construal-level theory of psychological distance},
  author={Trope, Yaacov and Liberman, Nira},
  journal={Psychol. Rev.},
  volume={117},
  number={2},
  pages={440-463},
  year={2010},
  publisher={American Psychological Association}
}

@article{trope2007construal,
  title={Construal levels and psychological distance: Effects on representation, prediction, evaluation, and behavior},
  author={Trope, Yaacov and Liberman, Nira and Wakslak, Cheryl},
  journal={J. Consum. Psychol.},
  volume={17},
  number={2},
  pages={83-95},
  year={2007},
  publisher={Elsevier}
}

@article{smyth2007recipient,
  title={Recipient syringe sharing and its relationship to social proximity, perception of risk and preparedness to share},
  author={Smyth, Bobby P and Roche, Aoife},
  journal={Addict. Behav.},
  volume={32},
  number={9},
  pages={1943-1948},
  year={2007},
  publisher={Elsevier}
}

@article{gattig2007judgmental,
  title={Judgmental discounting and environmental risk perception: Dimensional similarities, domain differences, and implications for sustainability},
  author={Gattig, Alexander and Hendrickx, Laurie},
  journal={J. Soc. Issues},
  volume={63},
  number={1},
  pages={21-39},
  year={2007},
  publisher={Wiley Online Library}
}

@article{she2012psychological,
title={Psychological Distance Model for Environmental Risk Communication},
author={She, Shengxiang and Ma, Chaoqun and Lu, Qiang and Xie, Xiaoling},
journal={Syst. Eng.},
volume={30},
number={9},
pages={69-74},
year={2012},
}

@book{lambert1970psychological,
  title={Psychological Stress and the Coping Process},
  author={Lazarus, R. S.},
  address={New York, NY, USA},
  publisher={McGraw-Hill},
  year={1966}
}

@article{kruglanski1996motivated,
title={Motivated closing of the mind: `seizing' and `freezing'},
author={Kruglanski, A. W. and Webster, D. M.},
journal={Psychol. Rev.},
volume={103},
number={2},
pages={263-283},
year={1996},
url={https://api.semanticscholar.org/CorpusID:19253040}
}

@article{kryvasheyeu2015performance,
  title={Performance of social network sensors during {Hurricane Sandy}},
  author={Kryvasheyeu, Yury and Chen, Haohui and Moro, Esteban and Van Hentenryck, Pascal and Cebrian, Manuel},
  journal={PLoS One},
  volume={10},
  number={2},
  pages={e0117288},
  year={2015},
  publisher={Public Library of Science San Francisco, CA USA}
}

@inproceedings{liu2025pychoagent,
    title = "{P}ycho{A}gent: Psychology-driven {LLM} Agents for Explainable Panic Prediction on Social Media during Sudden Disaster Events",
    author = "Liu, Mengzhu  and Zhu, Zhengqiu  and Ai, Chuan  and Gao, Chen  and Li, Xinghong  and others",
    booktitle = "Proc. Conf. Empir. Methods Nat. Lang. Process. (EMNLP)",
    month = nov,
    year = "2025",
    address = "Suzhou, China",
    publisher = "Association for Computational Linguistics",
    url = "https://aclanthology.org/2025.emnlp-main.865/",
    doi = "10.18653/v1/2025.emnlp-main.865",
    pages = "17116-17134",
    ISBN = "979-8-89176-332-6"
}

@article{bass1969new,
  title={A new product growth for model consumer durables},
  author={Bass, Frank M},
  journal={Manage. Sci.},
  volume={15},
  number={5},
  pages={215-227},
  year={1969},
  publisher={INFORMS}
}

@article{muslim2024mass,
  title={Mass media and its impact on opinion dynamics of the nonlinear q-voter model},
  author={Muslim, Roni and Nqz, Rinto Anugraha and Khalif, Muhammad Ardhi},
  journal={Phys. A: Stat. Mech. Appl.},
  volume={633},
  pages={129358},
  year={2024},
  publisher={Elsevier}
}

@inproceedings{yin2024agent,
  title={Agent-based modeling of {COVID-19} vaccine uptake in {New York} State: Information diffusion in hybrid spaces},
  author={Yin, Fuzhen and Jiang, Na and Crooks, Andrew and Laurian, Lucie},
  booktitle={Proc. 7th ACM SIGSPATIAL Int. Workshop GeoSpatial Simul.},
  pages={11-20},
  year={2024}
}

@article{rende2025crowd,
  title={Crowd: A Social Network Simulation Framework},
  author={Rende, Ann Nedime Nese and Yilmaz, Tolga and Ulusoy, Ozgur},
  journal={IEEE Trans. Comput. Soc. Syst.},
  year={2025},
  volume={12},
  number={6},
  pages={4369-4383},
  publisher={IEEE}
}

@article{wu2018modelling,
  title={Modelling Internet Users' Negative Emotion Based on {SOAR} Model},
  author={Wu, P and Qiang, SH and Gao, QN},
  journal={Chin. J. Manag. Sci.},
  volume={26},
  number={3},
  pages={126-138},
  year={2018}
}

@inproceedings{naskar2020predicting,
  title={Predicting emotion dynamics sequence on {T}witter via deep learning approach},
  author={Naskar, Debashis and Onaindia, Eva and Rebollo, Miguel and Singh, Sanasam Ranbir},
  booktitle={Proc. 18th Int. Conf. Adv. Mobile Comput. Multimedia (MoMM)},
  pages={20-24},
  year={2020}
}

@inproceedings{wang2024information,
  title={Information diffusion prediction with graph neural ordinary differential equation network},
  author={Wang, Ding and Zhou, Wei and Hu, Songiln},
  booktitle={Proc. 32nd ACM Int. Conf. Multimedia (ACM MM)},
  pages={9699--9708},
  year={2024}
}

@inproceedings{huang2024ecr,
title={{ECR-Chain}: Advancing Generative Language Models to Better Emotion-Cause Reasoners through Reasoning Chains},
author={Zhaopei Huang and Jinming Zhao and Qin Jin},
booktitle={Proc. 33rd Int. Joint Conf. Artif. Intell. (IJCAI)},
pages     = {6288-6296},
year      = {2024},
month     = {Aug.},
doi       = {10.24963/ijcai.2024/695},
}

@ARTICLE{zhang2024modeling,
  author={Zhang, Jiayi Eurus and Broekens, Joost and Jokinen, Jussi P. P.},
  journal={IEEE Trans. Affect. Comput.}, 
  title={Modeling Cognitive-Affective Processes With Appraisal and Reinforcement Learning}, 
  year={2025},
  volume={16},
  number={2},
  pages={771-782},
  doi={10.1109/TAFFC.2024.3470555}}

@article{piao2025agentsociety,
title={{AgentSociety}: Large-Scale Simulation of {LLM}-Driven Generative Agents Advances Understanding of Human Behaviors and Society},
author={Piao, Jinghua and  Yan, Yuwei and  Zhang, Jun and  Li, Nian and  Yan, Junbo and  Lan, Xiaochong and  Lu, Zhihong and  Zheng, Zhiheng and  Wang, Jing Yi and  Zhou, Di and  Gao, Chen and  Xu, Fengli and  Zhang, Fang and  Rong, Ke and  Su, Jun and  Li, Yong},
journal={arXiv:2502.08691},
year={2025},
}

@incollection{liberman2007psychological,
  author    = {Liberman, Nira and Trope, Yaacov and Stephan, Elena},
  title     = {Psychological distance},
  booktitle = {{Social Psychology: Handbook of Basic Principles}},
  edition   = {2nd},
  publisher = {Guilford Press},
  address   = {New York, NY, USA},
  pages     = {353-381},
  year      = {2007}
}

@inproceedings{xue2026some,
title={{SoMe}: A Realistic Benchmark for {LLM}-based Social Media Agents},
author={Xue, Dizhan and Cui, Jing and Qian, Shengsheng and Hu, Chuanrui and Xu, Changsheng},
booktitle={Proc. AAAI Conf. Artif. Intell. (AAAI)},
volume={40},
number={2},
pages={1391-1399},
year={2026}
}

@inproceedings{zhong2026modeling,
  title={Modeling Multi-Dimensional Cognitive States in Large Language Models under Cognitive Crowding},
  author={Zhong, Lin and Zhu, Siyu and Yuan, Zizhen and Cui, Jinhao and Zhao, Xinyang and Wang, Lingzhi and Chen, Hao and Liao, Qing},
  booktitle = "Findings Assoc. Comput. Linguist.: ACL 2026",
month = {Jul.},
  year = "2026",
  address = "San Diego, CA, USA",
  doi = "10.18653/v1/2026.findings-acl.1017",
  pages = "20322-20350",
}

@ARTICLE{popescu2014,
  author={Popescu, Alexandru and Broekens, Joost and van Someren, Maarten},
  journal={IEEE Trans. Affect. Comput.}, 
  title={{GAMYGDALA}: An Emotion Engine for Games}, 
  year={2014},
  volume={5},
  number={1},
  pages={32-44},
  doi={10.1109/T-AFFC.2013.24}}

@inproceedings{jiang2007ebdi,
title={{EBDI}: an architecture for emotional agents},
author={Hong Jiang and Jose M.Vidal and Michael N.Huhns},
booktitle={Proc. 6th Int. Joint Conf. Auton. Agents Multiagent Syst. (AAMAS)},
pages={1-3},
year={2007}
}

@inproceedings{NEURIPS2024_b0049c3f,
title={Apathetic or empathetic? evaluating {LLM}s' emotional alignments with humans},
author={Huang, Jen-tse and Man Ho Lam and Eric John Li and Shujie Ren and Wenxuan Wang and Wenxiang Jiao and Zhaopeng Tu and Michael R. Lyu},
booktitle={Proc. Adv. Neural Inf. Process. Syst. (NeurIPS)},
year={2024},
doi = {10.52202/079017-3077},
pages = {97053-97087},
url = {https://proceedings.neurips.cc/paper_files/paper/2024/file/b0049c3f9c53fb06f674ae66c2cf2376-Paper-Conference.pdf},
volume = {37}
}

@ARTICLE{Ala2025Aware,
  author={Tak, Ala N. and Gratch, Jonathan and Scherer, Klaus R.},
  journal={IEEE Trans. Affect. Comput.}, 
  title={Aware Yet Biased: Investigating Emotional Reasoning and Appraisal Bias in Large Language Models}, 
  year={2025},
  volume={16},
  number={4},
  pages={2871-2880},
  doi={10.1109/TAFFC.2025.3581461}}


\textbf{Mengzhu Liu} received the B.S. degree from Dalian Maritime University, Dalian, China, in 2023. She is currently a fourth-year Ph.D. student with the State Key Laboratory of Digital Intelligent Modeling and Simulation, National University of Defense Technology, Changsha, China. Her current research interests include cognition theory--driven affective computing, large language model--driven cognitive agents, and emergency decision-making simulation.

\textbf{Long Qin} received the BS degree in simulation engineering from the National University of Defense Technology, the MS and PhD degrees in control science and engineering from the National University of Defense Technology, and completed a six-month exchange study at the University of Leeds, UK. His research focuses on spatial reasoning, intelligent human-computer interaction, and complex system modeling and simulation, aiming to enable multimodal human-computer interaction experiences in virtual-real fusion simulation scenarios and to create realistic and trustworthy virtual character models. His broader interests include high-performance simulation platforms, deep reinforcement learning, and embodied vision-language-action modeling.

\textbf{Chuan Ai} received the PhD degree in Control Science and Engineering from the National University of Defense Technology, China, in 2021. His research focuses on modeling and simulation of social public opinion dissemination and social crowd behavior, covering opinion propagation mechanisms, crowd dynamics, and scenario-based simulation. His broader interests include high-performance simulation, large-scale simulation systems, and computational modeling for social systems.

\textbf{Zhengqiu Zhu} received the Ph.D. degree in management science and engineering from National University of Defense Technology, Changsha, China, in 2023. He is currently a associate professor with the College of Systems Engineering, National University of Defense Technology, Changsha, China. His research interests include embodied intelligence, crowd computing, and AI agents.  He serves as the Youth Editor of The Innovation. He also served as the Guest Editor of Pattern Recognition.

\textbf{Hongru Liang} received the PhD degree in computer science and technology from Nankai University. She is currently an associate professor at Sichuan University. Her research focuses on autonomous knowledge management for large language model-based agents, including multimodal knowledge representation, proactive knowledge acquisition, and efficient knowledge utilization. Her other research interests include self-evolving agents and computational musicology.

\textbf{Fangfang Li} is now a professor in the School of Computer Science and Engineering of Central South University. She received her M.S. and Ph.D. degrees in the School of Remote Sensing and Information Engineering, Wuhan University, Wuhan, China, in 2006 and 2009, respectively. Her research interests include natural language processing and text mining. She has presided over the National Natural Science Foundation Project of China and that of Hunan Province.

\textbf{Chen Gao} received the bachelor’s degree in electronic information science and technology and the Ph.D. degree in information and communication engineering from the Department of Electronic Engineering, Tsinghua University, Beijing, China, in 2021 and 2016, respectively. 
He is currently a Research Assistant Professor with Tsinghua University. His research primarily focuses on data mining and information retrieval, with over 100 papers in top-tier venues, attracting over 11000 citations. Dr. Gao’s work on GNN-based bundle recommendation received the Best Short Paper Honorable Mention Award in SIGIR 2020.

\textbf{Yong Li} (Senior Member, IEEE) received the B.S. degree in electronics and information engineering from Huazhong University of Science and Technology, Wuhan, China, in 2007, and the Ph.D. degree in electronic engineering from Tsinghua University, Beijing, China, in 2012. He is currently a Full Professor with the Department of Electronic Engineering, Tsinghua University. 

\textbf{Xin Lu} received the Ph.D. degree from Karolinska Institutet, Sweden. He is currently a Professor and the Chief Scientist with the College of Systems Engineering, National University of Defense Technology. He was a recipient of the National Science Fund for Distinguished Young Scholars, the Co-Founder and the Chief Analyst of the International Humanitarian Big Data Organization Flowminder Foundation, and the Deputy Director of Hunan Provincial Key Laboratory of Intelligent Decision Technology for Emergency Management. His long-term research focuses on big data mining, complex systems and networks, and emergency management.  

\textbf{Quanjun Yin} is currently a professor with the College of Systems Engineering, National University of Defense Technology, China. His research interests include cognitive process modelling, qualitative spatial reasoning and planning, cooperation and negotiation, cloud-based simulation, and edge computing.  

\clearpage
\appendix

\section{Appendix}
\label{sec:Appendix}
\subsection{PPDTS Questionnaire}
\label{subsec:ppdts_ques}

\noindent The Psychological Preparedness for Disaster Threat Scale (PPDTS)
comprises 18 items rated on a 4-point Likert scale (1 = completely
disagree, 4 = completely agree). Items are grouped into four factors;
reverse-scored items are marked with $(\dagger)$ and transformed via
$s' = 5 - s$ before factor aggregation. Item wording is reproduced
verbatim from the assessment instrument; the item-to-factor and
factor-to-desire mappings used in our model are summarized in
Table~\ref{tab:ppdts-mapping} of the main text.

\begin{enumerate}\setlength{\itemsep}{1pt}
\item I am familiar with the natural hazard/disaster preparedness
      materials relevant to my area $(\dagger)$
\item I know how to adequately prepare my home for the forthcoming
      fire/flood/cyclone season $(\dagger)$
\item I know which household preparedness measures are needed to stay
      safe in a natural hazard/disaster $(\dagger)$
\item I know what to look out for in my home and workplace if an
      emergency weather situation should develop
\item I am familiar with the disaster warning system messages used for
      extreme weather events $(\dagger)$
\item I am confident that I know what to do and what actions to take
      in a severe weather situation $(\dagger)$
\item I would be able to locate the natural hazard/disaster
      preparedness materials in a warning situation easily $(\dagger)$
\item I am knowledgeable about the impact that a natural
      hazard/disaster can have on my home
\item I know what the difference is between a disaster warning and a
      disaster watch situation
\item I am familiar with the weather signs of an approaching
      fire/flood/cyclone $(\dagger)$
\item I think I am able to manage my feelings pretty well in difficult
      and challenging situations
\item In a natural hazard/disaster situation I would be able to cope
      with my anxiety and fear
\item I seem to be able to stay cool and calm in most difficult
      situations
\item I feel reasonably confident in my own ability to deal with
      stressful situations that I might find myself in
\item When necessary, I can talk myself through challenging situations
\item If I found myself in a natural hazard/disaster situation I would
      know how to manage my own response to the situation
\item I know which strategies I could use to calm myself in a natural
      hazard/disaster situation
\item I have a good idea of how I would likely respond in an emergency
      situation
\end{enumerate}


\subsection{Prompt Templates}
\label{subsec:prompts}

\noindent Figure~\ref{box:system} and Figure~\ref{box:assessment}
reproduce the system-level and assessment prompts used to elicit
PPDTS responses from the role-playing LLM
(Qwen2.5-72B-Instruct). The two-shot answer-format constraint in the
assessment prompt enforces machine-parseable output
(\texttt{Q[n]: [score] (reason)}), from which the 18 scores are
extracted by regular expression.

\begin{figure}[t!]
\begin{tcolorbox}[%
  title={\textbf{System prompt for LLM role-playing}},
  fontupper=\scriptsize,
  colback=teal!8,
  colframe=teal!80!black,
  fonttitle=\bfseries\small,
  boxrule=0.6pt, arc=1.5mm]
You are role-playing as a Twitter user during Hurricane Sandy. You
must answer questions from this user's perspective, considering
their personality, experiences, and the current situation. Do NOT
mention that you are an AI or language model.
\end{tcolorbox}
\caption{System prompt for LLM role-playing.}
\label{box:system}
\end{figure}

\begin{figure}[t!]
\begin{tcolorbox}[%
  title={\textbf{Assessment prompt (PPDTS elicitation)}},
  label={box:assessment},
  fontupper=\scriptsize,
  colback=teal!8,
  colframe=teal!80!black,
  fonttitle=\bfseries\small,
  boxrule=0.6pt, arc=1.5mm]
\textbf{User prompt:}\\
Based on this user profile and current situation, answer the
following 18 questions as if you ARE this user during Hurricane
Sandy.\\[2pt]
\{user\_profile\_text\}\quad \{situation\_text\}\\[2pt]
Answer these questions from the user's perspective:\\
1.~I am familiar with the natural hazard/disaster preparedness
materials relevant to my area \quad\dots\quad 18.~I have a good idea
of how I would likely respond in an emergency situation\\[3pt]
\textbf{Response requirements:}
(1)~Answer ALL 18 questions sequentially; (2)~Format:
\texttt{Q[number]: [score] (reason)}, e.g.,
\texttt{Q1: 3 (I've seen hurricane materials before)};
(3)~Use the 1--4 rating scale: 1 = Completely Disagree, 2 = Somewhat Agree, 3 = Mostly Agree, 4 = Completely Agree; (4)~Do not include any
internal thinking process; (5)~Provide answers in direct format
without additional text.\\[3pt]
\textbf{Model output (excerpt):}\\
Q1: 2 (I have heard of preparedness kits but never assembled one)\\
Q2: 3 (I know the basics of boarding up windows)\quad\dots
\end{tcolorbox}
\caption{Assessment prompt (PPDTS elicitation)}
\label{box:assessment}
\end{figure}


\subsection{Implementation Parameters}
\label{subsec:params}

This appendix lists the dataset- and implementation-specific
parameters that, while not part of the theoretical model, are
required for full reproducibility. All values are taken from the
reference implementation released at
\url{https://github.com/supersonic0919/PanicCognitivePath}.

\subsubsection{Hurricane Category to Loss Magnitude ($x$)}

The loss magnitude $x$ used in the PSD computation is mapped from
the hurricane category as follows. We use a linear scale with unit
step 100 so that $x$ spans a six-fold range across the
Saffir--Simpson scale, reflecting the non-linear damage escalation
reported in post-event surveys.

\begin{table}[h]
\caption{Category-to-loss mapping.}
\label{tab:cat-loss}
\centering
\renewcommand{\arraystretch}{1.10}
\begin{tabular*}{\columnwidth}{@{\extracolsep{\fill}}l l r@{}}
\toprule
\textbf{Category} & \textbf{Name} & \textbf{\(x\)} \\
\midrule
D  & Tropical Depression & 100 \\
S  & Tropical Storm     & 200 \\
E  & Extratropical      & 300 \\
H1 & Category~1         & 400 \\
H2 & Category~2         & 500 \\
H3 & Category~3         & 600 \\
\bottomrule
\end{tabular*}
\end{table}

\subsubsection{Post-Dissipation Loss Decay}

After dissipation, we retain a non-zero $x$ to model the prolonged
impact of secondary hazards (flooding, power outages, logistics
disruption). The decay schedule was calibrated on the Sandy dataset
to match the empirical post-event tweet volume.
\begin{itemize}\setlength{\itemsep}{2pt}
\item \textbf{0--3 days after dissipation:} $x = 300$ (constant,
  equivalent to extratropical level).
\item \textbf{4--8 days after dissipation:}
  $x = 300 \cdot 0.92^{\,d-3}$ (slow exponential decay).
\item \textbf{$> 8$ days after dissipation:}
  $x = \max\!\big(100,\; 300 \cdot 0.92^{5} \cdot 0.85^{\,d-8}\big)$
  (fast exponential decay, floored at tropical-depression level).
\end{itemize}

\subsubsection{Spatial Distance Under Dissipation}

When the hurricane has dissipated, we cannot obtain a current storm
center. Instead, we use the geodesic distance from each user to the
last-known storm position and apply a dissipation decay factor
$D_d = 0.8$ to model the user's diminished spatial awareness after
the storm has passed. Users without geocoded locations fall back to
a fixed distance of $5000$~km (effectively excluded from the PSD
gate).

\subsubsection{Panic Exposure Dynamics}

The social panic exposure $n$ follows three regimes:
\begin{itemize}\setlength{\itemsep}{2pt}
\item \textbf{Normal regime:}
  $n = \frac{|\{f \in F_u : f \text{ panicked before step } t\}|}{|F_u|}$,
  where $F_u$ is the set of users $u$ follows.
\item \textbf{Cool-down regime (within 3 steps after $u$'s first panic):}
  $n = 0.99$ (locked near the maximum to model panic inertia).
\item \textbf{Recovery regime (3--13 steps after first panic):}
  $n$ linearly interpolates from $0.99$ back to its normal-regime
  value over 10 steps.
\end{itemize}
\noindent Users without follow data are assigned $n = 0.0$ (no panic-expressing agents in the attention list, hence no social amplification). $n$ is clamped to $[0.00, 0.99]$ in all regimes.

\subsubsection{LLM Configuration}

All role-playing calls use \texttt{Qwen2.5-72B-Instruct} served via
an OpenAI-compatible endpoint, with deterministic parsing by regular
expression and up to 3 retries on malformed responses. Per-agent
inputs comprise the pre-generated user profile text (personality,
experiences) and a per-step situation text (disaster state, spatial
and social context). The distance weights $a=b=c=1$ and loss
magnitude $p=1$ follow construal-level theory, under which the four
distance dimensions share a common psychological meaning and produce
similar effects on risk perception; no
domain-specific prior justifies differential weighting.

\subsection{Lexicon Tables}
\label{subsec:lexicon}

\noindent\textbf{Design principles.}
The lexicon partitions surface cues on the first panic-related tweet
of each delayed-panic user into four mutually exclusive categories.
Tokens are matched by case-insensitive whole-word regex with optional
part-of-speech tolerance; consequence-bearing tokens are deliberately
\emph{not} allowed in \textit{Primary} even when they co-occur with
event-identity tokens, in which case the \emph{Secondary}-overrides-
\emph{Primary} priority rule routes the tweet to \textit{Secondary}.

\subsubsection{Primary Patterns (Event Identity)}
\noindent Restricted to storm identity, physical parameters, weather
phenomena, and pure narrative cues.
\textbf{Consequence-bearing} tokens (e.g., devastate,
snow, blizzard, ravage, mandatory
evacuation, warning\slash watch\slash advisory) are explicitly excluded
and routed to Secondary.

\begin{itemize}\setlength{\itemsep}{2pt}\setlength{\parskip}{0pt}
\raggedright
\item \textbf{Hurricane naming / type:} sandy, hurricane, typhoon,
      cyclone, hurricane sandy, hurricanesandy, storm sandy,
      tropical storm(s), nor'easter, frankenstorm, superstorm.
\item \textbf{Physical parameter:} landfall, eye of (the) storm,
      storm surge (excl.\ consequence context),
      category 1--5 \slash\ cat 1--5.
\item \textbf{Weather phenomenon:} wind(s), gust(s)
      (excl.\ damage context), mph, high wind, wind gust,
      rain(ing), heavy rain.
\item \textbf{Pure narrative:} scary, frightened, terrified,
      horrified, panic(king) (excl.\ consequence context),
      frightening, terrifying, nightmarish,
      devastating (excl.\ havoc \slash\ damage context),
      reeling, tragedy, catastroph(e\slash ic), chaos,
      mayhem, wrath.
\end{itemize}
\subsubsection{Secondary Patterns (Consequence/Response)}
\noindent All consequence, response, infrastructure-failure, and
emergency-response cues.

\begin{itemize}\setlength{\itemsep}{2pt}\setlength{\parskip}{0pt}
\raggedright
\item \textbf{Physical harm:} damage(d\slash s), destroy(ed),
      demolish(ed), wreck(ed\slash age), destruction,
      devastate(d\slash ion\slash e\slash es), ravage(d\slash s),
      havoc, debris, aftershock, victim(s), suffer(ed\slash ing), mold.
\item \textbf{Utility outage:} power\slash electricity
      outage\slash out\slash cut off\slash lost\slash fail,
      no power\slash electricity\slash water\slash gas\slash heat,
      without (same), blackout(s), outage(s), power line(s).
\item \textbf{Flooding:} flood(ing\slash ed\slash s\slash water),
      groundwater, stormwater, water rising.
\item \textbf{Snow / ice (Sandy-stage sequels):}
      snow(storm)s?, blizzard, ice storm, freezing.
\item \textbf{Evacuation/shelter:} evacuate (any form),
      mandatory evacuation,
      (hurricane\slash tropical)storm warning\slash watch\slash advisory,
      shelter(s\slash ed\slash ing), refus(e\slash ed\slash ing),
      displaced, stranded, trapped,
      stay inside\slash indoors\slash home\slash away.
\item \textbf{Casualties:} death toll,
      died\slash dead\slash killed\slash kill(ing),
      casualty(ies), injur(y\slash ied\slash ies), missing.
\item \textbf{Emergency response / recovery:}
      rescue(d\slash rs), first responder, fema, red cross,
      national guard, insurance, claim(s\slash ed), relief, aid,
      donat(e\slash ed\slash s), volunteer(s\slash ed\slash ing),
      clean(up), recovery, rebuild(ing).
\item \textbf{Resource shortages / sanitation:} shortage(s),
      running out, empty shelves,
      fuel\slash gas shortage, contamination, toxic, sewage.
\item \textbf{Follow-on hazards/crime:} fire
      (with start\slash burn\slash outbreak), looting,
      crime (spree\slash wave\slash surge\slash rise\slash rate),
      gas leak.
\end{itemize}

\subsubsection{Social Patterns (Concern for Others)}
\noindent Reserved for cues expressing other-directed affect
(family, relatives, neighbors, prayer, information relay).
Pure fear tokens such as afraid are routed here when used
with (for\slash about\slash of) \{others\}.

\begin{itemize}\setlength{\itemsep}{2pt}\setlength{\parskip}{0pt}
\raggedright
\item \textbf{Kinship\slash community:} my (mom\slash dad\slash family\slash
      friends\slash brother\slash sister\slash kids\slash parents\slash cousin\slash
      uncle\slash aunt\slash ...), our (neighbors\slash ...),
      his\slash her (family\slash ...).
\item \textbf{Affective\slash prayerful:} praying, thoughts and prayers,
      stay safe, please (be safe\slash ...\slash donate).
\item \textbf{Check-in\slash worry:} check (on\slash in),
      (scared\slash afraid) (for\slash about\slash of), worried about,
      people\slash those\slash anyone (are\slash is) (affected\slash ...).
\item \textbf{Information relay:} watching (the) news,
      saw (on) news, heard from,
      saw on (cnn\slash bbc\slash ...),
      my thoughts (go\slash are\slash with).
\end{itemize}

\subsubsection{Priority Rule}
\noindent Categories are assigned in the priority order
\textbf{Secondary $>$ Primary $>$ Social $>$ Other}; a tweet is
assigned to the highest-priority category whose lexicon yields at
least one match on the tweet body. Ties at the same priority level
are broken by token count (the matching category with the greater
number of distinct trigger tokens wins); residual ties fall back
to Other.

\end{document}